\documentclass[nohyperref]{article}

\usepackage{microtype}
\usepackage{graphicx}
\usepackage{subfigure}
\usepackage{booktabs} % for professional tables

\usepackage{hyperref}

\usepackage[accepted]{icml2023}

\usepackage{amsmath}
\usepackage{amssymb}
\usepackage{mathtools}
\usepackage{amsthm}
\usepackage{multirow}

\usepackage[capitalize,noabbrev]{cleveref}

\newcommand{\showcomments}{yes}
\newcommand\xingjian[1]{
\ifthenelse{\equal{\showcomments}{yes}}{{\color{red} XJ: #1}}{\ignorespaces}
}
\newcommand\bingzhao[1]{
\ifthenelse{\equal{\showcomments}{yes}}{{\color{blue} BZ: #1}}{\ignorespaces}
}
\newcommand\nick[1]{
\ifthenelse{\equal{\showcomments}{yes}}{{\color{green} Nick: #1}}{\ignorespaces}
}

\theoremstyle{plain}

\theoremstyle{definition}

\theoremstyle{remark}

\usepackage[textsize=tiny]{todonotes}

\icmltitlerunning{Submission and Formatting Instructions for ICML 2023}

\begin{document}

\twocolumn[
\icmltitle{XTab: Cross-table Pretraining for Tabular Transformers}



\icmlsetsymbol{5}{*}
\icmlsetsymbol{6}{\textdagger}

\begin{icmlauthorlist}
\icmlauthor{Bingzhao Zhu}{1,2,5}
\icmlauthor{Xingjian Shi}{3,6}
\icmlauthor{Nick Erickson}{4}
\icmlauthor{Mu Li}{3,6}
\icmlauthor{George Karypis}{4}
\icmlauthor{Mahsa Shoaran}{1}
\end{icmlauthorlist}

\icmlaffiliation{1}{EPFL, Lausanne, Switzerland}
\icmlaffiliation{2}{Cornell University, Ithaca, USA}
\icmlaffiliation{3}{Boson AI, USA}
\icmlaffiliation{4}{Amazon Web Services, USA}

\icmlcorrespondingauthor{Bingzhao Zhu}{bz323@cornell.edu}

\icmlkeywords{Machine Learning, ICML}

\vskip 0.3in
]



\printAffiliationsAndNotice{$^{*}$Work done as an intern at Amazon Web Services. $^{\dagger}$Work done while being at Amazon Web Services.}  % leave blank if no need to mention equal contribution

\begin{abstract}
The success of self-supervised learning in computer vision and natural language processing has motivated pretraining methods on tabular data. However, most existing tabular self-supervised learning models fail to leverage information across multiple data tables and cannot generalize to new tables. %\xingjian{Consider to revise the sentence to focus on the contribution of XTab, i.e. leverage cross-table information.} 
In this work, we introduce XTab, a framework for cross-table pretraining of tabular transformers on datasets from various domains. We address the challenge of inconsistent column types and quantities among tables by utilizing independent featurizers and using federated learning to pretrain the shared component. Tested on 84 tabular prediction tasks from the OpenML-AutoML Benchmark (AMLB), we show that (1) XTab consistently boosts the generalizability, learning speed, and performance of multiple tabular transformers, (2) by pretraining FT-Transformer via XTab, we achieve superior performance than other state-of-the-art tabular deep learning models on various tasks such as regression, binary, and multiclass classification.
\end{abstract}

\vspace{-0.3in}
\section{Introduction} \vspace{-0.05in}
\label{introduction}
With the increasing number of datasets represented as tables with rows and columns, tabular machine learning makes the foundation of many real-world applications. While deep learning has achieved tremendous success in the fields of computer vision (CV) \citep{he2022masked, liu2021swin} and natural language processing (NLP) \citep{devlin2018bert, vaswani2017attention}, tabular deep learning models are not used as commonly as tree-based models~\citep{grinsztajn2022tree, gijsbers2022amlb}. 
%in the area of tabular data analysis. 
The primary challenge of tabular deep learning is the diversity of tabular tasks. Unlike text, which can be standardized as a sequence of tokens, tables are highly data-specific. Tabular data can vary in the number and types of columns. This makes it difficult for tabular deep learning models to transfer the knowledge learned from one table to another, leading to poor generalization abilities. Therefore, self-supervised learning for tabular data~\citep{he2022masked, devlin2018bert}, particularly one that is able to bootstrap the learning on new tables, is still an open problem.

%\citep{wang2022transtab}. Unlike CV and NLP \citep{he2022masked, devlin2018bert}, a foundation tabular model has not yet been proposed.

There is an ongoing effort in migrating self-supervised pretraining techniques from CV~\citep{chen2020simple} and NLP~\citep{devlin2018bert} to tabular tasks.
%For instance, self-supervised learning was proposed to alleviate label sparsity in NLP~\citep{devlin2018bert} and CV.
With self-supervised pretraining, tabular deep models have demonstrated improved performance~\citep{ucar2021subtab, bahri2021scarf, majmundar2022met}. However, existing methods generally pretrain the tabular model on data from the same domain as the downstream task. As a result, the data-specific models cannot generalize to new tables.
%As a result, the pretraining data is often limited to the training set of each downstream task~\citep{bahri2021scarf}.
%, which makes these self-supervised learning method task-specific.
%In other words, tabular self-supervised learning is performed in a task-specific manner, and is not re-used to improve performance on future tasks. % the pretrained backbone fails to work on new tabular tasks.

Another direction of deep tabular learning aims to leverage Transformers, which drives the recent progress in NLP~\citep{vaswani2017attention} and CV~\citep{dosovitskiy2020image} for tabular tasks. Inspired by the success of the attention mechanism, 
Transformers were adapted to tabular data~\citep{gorishniy2021revisiting, somepalli2021saint, wu2021fastformer, wang2022transtab} and demonstrated strong performance~\citep{grinsztajn2022tree}. The core idea of tabular transformers is to consider the table columns as tokens, similar to words in a sentence. Therefore, tabular transformers can process tables with variable numbers of columns, thus making transferable learning \citep{wang2022transtab} feasible.

In this paper, we present \textit{XTab}, a general framework for \textit{cross-table pretraining of tabular transformers}.
%XTab pretrains tabular Transformers on a large collection of tabular datasets. 
To resolve the issue that tables may vary in the number and types of columns, XTab decomposed the tabular transformers to two components: data-specific featurization and projection layers that capture the characteristics of each table, and a cross-table-shared block that stores the common knowledge. On a diverse collection of data tables, XTab trains these data-specific blocks and the shared block jointly via federated learning~\cite{collins2022fedavg}. Once pretrained, XTab can bootstrap the learning process on a new table by initializing the shared block with pretrained weights.  To verify our design, we conducted extensive experiments on AutoML Benchmark (AMLB)~\cite{gijsbers2022amlb}. Our results show that transformers pretrained and initialized with XTab consistently outperform
transformers with random initialization. By pretraining FT-Transformer~\citep{gorishniy2021revisiting} with XTab, we outperform the state-of-the-art tabular deep learning models.

%the above-mentioned recent advancement in tabular deep learning, i.e a Transformer architecture and self-supervised pretraining, to build a foundation model for various tabular tasks. 
The contributions of the paper are summarized as follows:
\begin{itemize} \vspace{-0.1in}
\itemsep-0.3em 
\item XTab offers a framework to account for cross-table variations and enable cross-table knowledge transfer. % allows us to pretrain a Transformer on multiple tabular tasks.  
\item Given the large diversity of tabular datasets, we propose to pretrain on tabular datasets with federated learning. This allows us to perform distributed pretraining across a large collection of tables. %large-scale pretraining.
\item To the best of our knowledge, we are the first to show that cross-table pretraining can boost the learning speed and performance on new tables. This is different from table understanding tasks~\cite{yin2020tabert},  the focus of which is to extract the semantical information from tables.
%help learn a shar builds the first foundation model that can \emph{generalize across tables}. Testing on the AutoML Benchmark, we found that Transformers pretrained and initialized with XTab outperform Transformers with random initialization on various downstream tasks.
\end{itemize}
\vspace{-0.2in}
\section{Related work} \label{related_work} \vspace{-0.05in}
\paragraph{Tabular self-supervised learning.} Inspired by the success of pretraining in CV and NLP, previous papers studied tabular self-supervised learning \citep{yoon2020vime, ucar2021subtab, somepalli2021saint, bahri2021scarf, majmundar2022met, rubachev2022revisiting, wang2022transtab}. Among those works, \citet{yoon2020vime, ucar2021subtab} proposed an auto-encoder framework with a pretext task to reconstruct the missing part of a table. \citet{bahri2021scarf} used contrastive learning as the pretraining objective and extended the SimCLR framework~\citep{chen2020simple} to tabular tasks. \citet{rubachev2022revisiting, wang2022transtab} further incorporated the label columns of tabular tasks in pretraining and proposed ``target-aware" objectives leading to higher performance. As existing approaches only pretrain on one \citep{bahri2021scarf,ucar2021subtab} or a few relevant tables \citep{wang2022transtab}, the pretrained tabular model lacks generalizability. XTab alleviates this issue by pretraining on a large number of tables. %\xingjian{TODO: Try to highlight the novelty of XTab.}
\vspace{-0.15in}
\paragraph{Tabular transformers.} Transformer models are gaining popularity in the realm of deep learning for tabular data. For example,
FT-Transformer has demonstrated superior performance on tabular classification/regression tasks~\citep{gorishniy2021revisiting}. Saint introduces the row-wise attention and captures the inter-sample interactions using transformer~\citep{somepalli2021saint}. Fastformer proposes to use additive attention on tabular tasks, which is a lightweight attention mechanism with linear complexity to the length of input sequences~\citep{wu2021fastformer}. TransTab features transfer learning in tabular tasks using transformers~\citep{wang2022transtab} and also supports the cross-table transfer. Our approach is different from TransTab in that TransTab has limited ability in generalizing to tables from new domains, while XTab is able to generalize to new domains.

\paragraph{Cross-table transfer learning.} Pretrained vision and text models can be adapted to a wide range of tasks~\cite{bommasani2021opportunities}. One reason is that the sentences and images share general representations across various tasks. 
As for tabular learning, one may question if there is shared knowledge across tables as two different tables can have totally different numbers of columns and the associated semantic meanings. We argue that different tables share a similar prior given the recent success of zero-shot hyperparameter optimization (HPO) in AutoML~\citep{winkelmolen2020practical}, which learns a general hyperparameter configuration applicable to a wide range of tabular tasks. Unlike pretrained models in NLP~\citep{devlin2018bert}, XTab does not attempt to learn a universal tokenizer for all tables, as the meaning and context of each table varies. Instead, we aim to learn a weight initialization that is generalizable to various downstream tasks. Concurrent to our work, tabular prior-data fitted networks~(TabPFN)~\citep{hollmann2022tabpfn} learns a prior model on synthetic tabular data and demonstrated promising results on small numerical tabular classification tasks with $\leq1000$ samples. Different from TabPFN, the inference complexity of XTab is irrelevant to the number of training samples. Thus, XTab also works for large tables.

\vspace{-0.1in}
\section{Methods} \vspace{-0.05in}
Previous works have proposed various pretraining methods for tabular learning~\citep{bahri2021scarf, ucar2021subtab, rubachev2022revisiting, somepalli2021saint}. However, existing pretrained models are still domain-specific since they were pretrained on the training set of each individual tabular prediction task. As a result, existing pretrained models lack generalizability and fail to cover downstream tasks on other types of tables. Here, we propose XTab to pretrain transformer models using the information from multiple tables. With cross-table pretraining, XTab aims to learn the shareable knowledge that can boost the performance for various downstream regression and classification tasks.

\begin{figure}[ht]
\vskip 0in
\begin{center}
\centerline{\includegraphics[width=0.85\columnwidth]{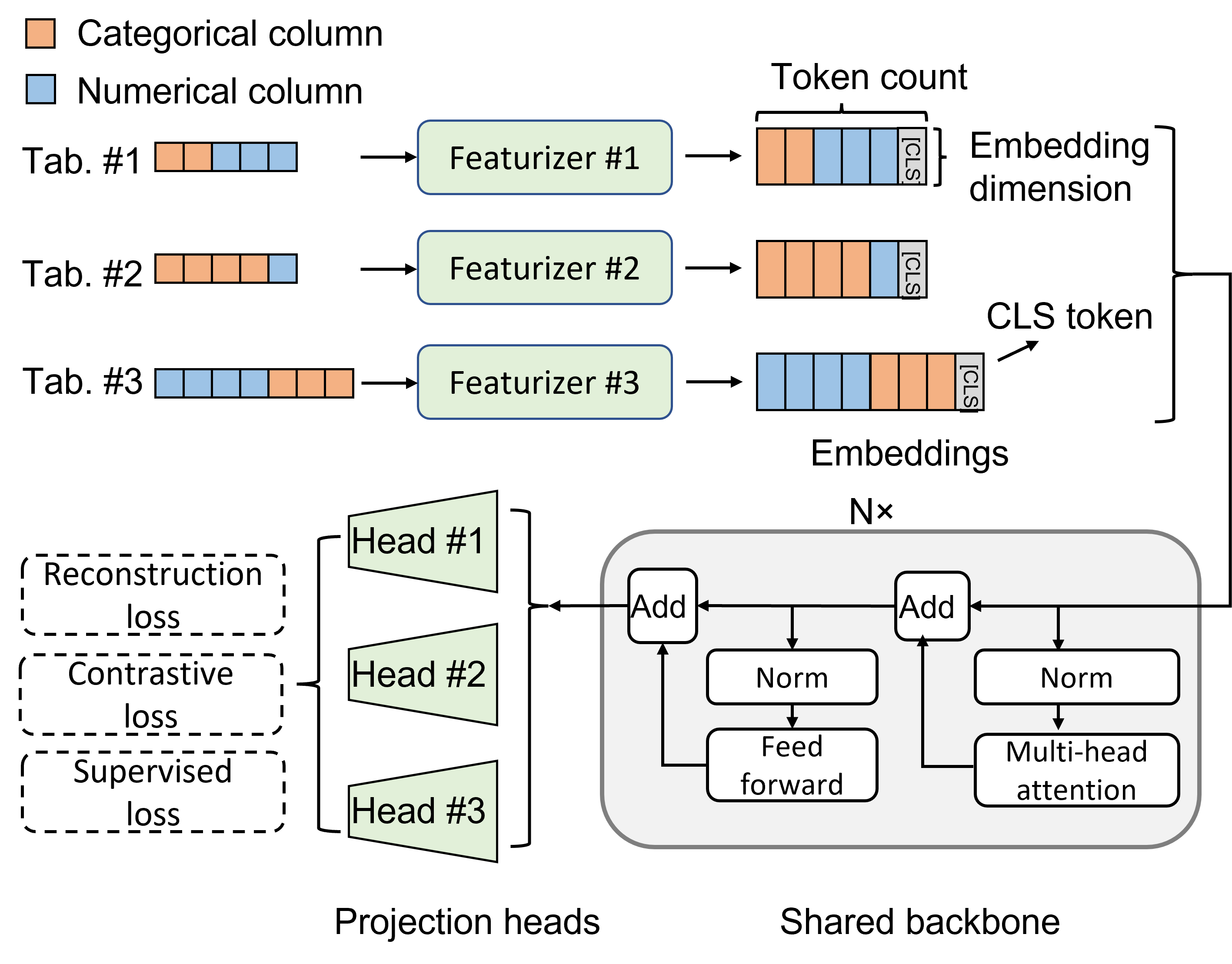}}
\vspace{-0.1in}
\caption{The model structure of XTab. XTab is pretrained on multiple tabular tasks (Tab. \#1, \#2, \#3). Samples from different tables are featurized and fed into a transformer model with N blocks. The output of the transformer is further processed by projection heads to derive the pretraining losses. Featurizers and projection heads are data-specific since tables may have different input/output dimensions. The transformer backbone is shared across all pretraining tables to capture the general knowledge.}
\label{model_structure}
\end{center}
%\vspace{-4.5em}
\vskip -0.4in
\end{figure}

\vspace{-0.1in}
\subsection{Model structure} \vspace{-0.05in}
The model structure of XTab is described in Figure \ref{model_structure}. During the pretraining phase, we sample mini-batches of rows from different tables (one batch per table). The featurizers are data-specific and convert each column of the table to a token embedding. An additional $\textsf{[CLS]}$ token is appended during this step for supervised prediction or contrastive self-supervised pretraining \citep{wang2022transtab}. A transformer-based backbone is shared across all tabular datasets to process token embeddings with variable sequence lengths. The output of the shared backbone is further processed by projection heads to (1) reconstruct the original table from a corrupted view; (2) identify the positive/negative pairs of samples as in contrastive learning; or (3) predict the values in the label column predefined by each table. The projection heads are not shared across tables since they are specific to each dataset and the pretraining objectives. Among all  pretraining losses, reconstruction loss and contrastive loss do not require information from the label column, whereas supervised losses use the groundtruth data in the label columns of each table. Using groundtruth information during the pretraining phase is referred to as ``target-aware pretraining"~\citep{rubachev2022revisiting, wang2022transtab} or ``pre-finetuning" \citep{aghajanyan2021muppet} in previous works.

A key challenge in cross-table pretraining lies in the variations of input tables. Previous works on transferable tabular learning either require tables to come from similar domains \citep{levin2022transfer} or use additional information (e.g., column names) to identify the shared knowledge across tables. XTab is designed to be applicable to previously unseen tables with no assumption on the domain or column name format. %XTab is designed to work with arbitrary tables with no additional information. xingjian{Do you mean "XTab is designed to be applicable to any tables with no assumption on the domain or column name format."?} 
To this end, XTab contains  model blocks that carry the data-specific information (green blocks in Figure~\ref{model_structure}), as well as the shared backbone that stores the common knowledge (grey blocks in Figure~\ref{model_structure}). Once pretrained, only a shared backbone is kept for all downstream tasks. For each downstream task, featurizers and projection heads are randomly initialized and the entire model is finetuned on the downstream training data until a stopping criterion is met.

\vspace{-0.1in}
\subsubsection{Featurizers} \label{Featurizers} \vspace{-0.05in}
The featurizers convert a sample to feature embeddings $E \in \mathbb{R}^{c \times d}$. Here, $c$ denotes the number of columns and $d$ is the embedding dimension. Each row of a table is considered as an input sample, and each column is a token. The embedding of $\textsf{[CLS]}$ token is appended to the feature embedding for prediction $\text{stack}[E, \textsf{[CLS]}] \in \mathbb{R}^{c+1 \times d}$. In this work, we limit our discussion to tables with numerical and categorical columns. Text cells are treated as categorical attributes. Our tokenizer is similar to \citet{gorishniy2021revisiting}. For numerical features, we multiply the numerical value $x_{k}$ at the $k$-th column with a trainable vector $W_k \in \mathbb{R}^{d}$ and add a bias term $b_k$. For categorical columns, XTab learns an embedding matrix $\in \mathbb{R}^{N_{cat} \times d}$ as a lookup table, where $N_{cat}$ is the total number of categories of the dataset. During the forward pass, we retrieve the categorical feature embeddings from the embedding matrix.

XTab allows tables to have different numbers of columns and arbitrary column types. Featurizers are data-specific to handle various types and numbers of columns in the input.

\vspace{-0.1in}
\subsubsection{Backbones} \label{method_backbone} \vspace{-0.05in}
As the shared component across multiple pretraining datasets, transformers can handle input sequences with variable lengths. Therefore, it is possible to pretrain a tabular transformer that can be applied to all tabular datasets. Compared with other deep learning architectures like multi-layer perceptron (MLP), transformers are  favorable for cross-table knowledge transfer since they can handle variable input sequences~\citep{wang2022transtab}. As long as the backbone can process input sequences of variable lengths, XTab is flexible on the exact implementation. In this work, we present three  backbone variants: 

\textbf{FT-Transformer:} Feature Tokenizer Transformer (FT-Transformer) is a simple yet well-performing transformer model for tabular prediction tasks \citep{gorishniy2021revisiting}. The transformer module in FT-Transformer consists of a Multi-Head Self-Attention (MHSA) block and a Feed Forward block \citep{vaswani2017attention}. Recent work has found FT-Transformers to beat other deep learning methods on tabular data \citep{grinsztajn2022tree}.

\textbf{Fastfromer:} Conventional Transformer-like architectures have a quadratic complexity to the length of input sequence \citep{vaswani2017attention}, making them inefficient for tables with large numbers of columns. Fastfromer is an efficient transformer architecture which uses additive attention in place of MHSA \citep{wu2021fastformer}. With additive attention, Fastformer only considers the interaction between each token and the global representation, achieving a linear complexity. 
%for tabular prediction tasks. 

\textbf{Saint-v:} Saint has introduced the row-wise attention in addition to the column-wise attention of FT-Transformer and Fastformer \citep{somepalli2021saint}. The original implementation of Saint is sensitive to the sequence length and can not handle variable-column tables \citep{somepalli2021saint}. We present a variation of Saint (Saint-v) to fit into our cross-table pretraining setting. Saint-v consists of both column- and row-wise attention blocks, and the detailed model structure is depicted in Appendix \ref{app_saint_v}. %\xingjian{List the section id of the appendix}.

% \noindent Detailed configurations of the Transformer backbones can be found in the Appendix.
\vspace{-0.1in}
\subsubsection{Projection heads and objectives} \label{obj} \vspace{-0.05in}
There exist various pretraining objectives for tabular prediction tasks \citep{rubachev2022revisiting, majmundar2022met, bahri2021scarf, ucar2021subtab, wang2022transtab, yoon2020vime}. Among them, table reconstruction and contrastive learning are the most popular and effective objectives for tabular tasks. In addition to the self-supervised pretraining objectives, we also tested the pre-finetuning setting using supervised loss.

\textbf{Reconstruction loss:} Reconstruction loss is a self-supervised training objective shown to be effective on various tabular tasks \citep{rubachev2022revisiting, majmundar2022met}. The reconstruction objective aims to recover the original sample $x$ from a corrupted view of the sample $\tilde{x}$. The reconstruction projection head takes the representation of $\tilde{x}$ as input, and generates an estimate of the original input $\hat{x}$. The reconstruction loss is calculated by comparing $x$ and $\hat{x}$. Specifically, we use Cross-Entropy loss to measure the reconstruction error of categorical columns and Mean Squared Error (MSE) for numerical columns.

\textbf{Contrastive loss:} Similar to the reconstruction objective, we also generate $\tilde{x}$ as a corrupted sample. $x$ and its corresponding corruption $\tilde{x}$ are considered as a positive pair of samples, whereas $x$ and other samples in the batch form negative sample pairs. In general, contrastive loss aims to minimize the distance between positive pairs of samples and maximize the distance for negative pairs. Following \citet{bahri2021scarf, chen2020simple}, we used InfoNCE loss for contrastive cross-table pretraining. The contrastive projection heads are similar to those used in SimCLR \citep{chen2020simple}, mapping the representations to the space where we apply the contrastive loss.

\textbf{Supervised loss:} In addition to reconstruction and contrastive losses that do not require labels in pretraining, one can directly pretrain a model using the supervised objective. With supervised losses, the projection head aims to predict the values under a certain field (or column), as predefined by each dataset. The supervised prediction tasks included regression and classification. 

In XTab, the projection heads are data-specific. Different pretraining datasets do not need to share common objectives. For example, we can simultaneously pretrain XTab on both regression and classification tasks, or a mixture of reconstruction and contrastive losses. The diversity of pretraining objectives ensures that the shared backbone is widely adaptable to various downstream tables.

\vspace{-0.1in}
\subsection{Federated pretraining} \label{federated_pretrain} \vspace{-0.05in}
XTab introduces data-specific featurizers and projection heads (green blocks in Figure \ref{model_structure}) to account for the variations across table columns and pretraining objectives. During pretraining, both the time and space complexity increase linearly as we include more tabular datasets. As a result, it is challenging to quickly pretrain XTab  using a single machine on a large collection of tabular tasks. 
To alleviate this issue, we fit XTab into the federated learning framework \citep{mcmahan2017communication}. With the federated setting, XTab involves only marginal overhead in wall-clock time with more pretraining tasks. Federated learning makes it feasible to pretrain XTab on a cluster of commercially available GPUs (NVIDIA T4 GPUs, 16GB memory). 

We use the Federated Averaging (FedAvg) algorithm to pretrain XTab \citep{mcmahan2017communication, li2019convergence}. We have a central server and multiple clients. Each client only hosts one dataset. Therefore, we can distribute the data-specific components of XTab across clients such that each client stores one featurizer, one projection head, and the shared transformer. During pretraining, each client calculates the gradient using the local dataset:
\begin{equation}
\label{eq1}
    w_{k, i+1} \leftarrow w_{k, i}- \alpha \nabla \ell_{k},
\end{equation}
\noindent where $k$ denotes the client (or table) index and $i$ shows the current iteration. $\alpha$ is the learning rate and $\ell^{(k)}$ is the loss function. $w$ represents the trainable parameters which contains two components: $w^{(\text{S})}$ for the shareable modules across all pretraining tasks, and $w^{(\text{NS})}$ for the non-shareable parts ($w  = \text{stack}[w^{(\text{NS})}, w^{(\text{S})}]$). All clients operate synchronously during pretraining with the same learning rate and batch size. 

The central server is responsible for aggregating the local gradients from clients. FedAvg allows clients to make multiple local updates before an aggregation step is made on the central server. Let $N$ denote the number of local updates per aggregation. The central server performs:
\begin{equation}
\label{eq2}
    w^{(\text{S})}_{i+N} \leftarrow w^{(\text{S})}_{i} +  \sum_{k=1}^K (w^{(\text{S})}_{k, i+N} - w^{(\text{S})}_{i}).
\end{equation}
The aggregation is only performed on the shared weights. The term $w^{(\text{S})}_{k, i+N} - w^{(\text{S})}_{i}$ is the gradient learned by client $k$ since the last weight aggregation. The central server simply accumulates the gradients from all clients. Such unitary scalarization was recently shown to perform well in multi-task learning \citep{kurin2022defense}. 

After the aggregation update (i.e., Equation \ref{eq2}), all clients download $w^{(\text{S})}_{i+N}$ from the central server, and apply the weights to the transformer backbone $w_{k, i+N}  = \text{stack}[w_{k, i+N}^{(\text{NS})}, w^{(\text{S})}_{i+N}]$. Therefore, we force all clients to train on a shared backbone with data-specific featurizers and projection heads.

The number of local steps $N$ is a key parameter to control communication efficiency. With $N=1$, FedAvg corresponds to the distributed version of stochastic gradient descent (SGD). With $N>1$, multiple local updates are performed between model aggregation steps at the server, thereby reducing the communication cost between the central server and clients.
Unless otherwise specified, we choose $N=5$ throughout the paper. The ablation study on $N$ is shown in Figure \ref{FedAvg} of the Appendix. % \xingjian{Are there any ablation analysis on the N? If so, it may also be good to put into Appendix.} % We test different choices of $N$ in the Experiments sessions.  % Previous works have shown that $N>1$ leads to improved representation learning \citep{collins2022fedavg}. 

Federated learning was originally proposed as a privacy-preserving approach to learning from distributed data. The collaboration of multiple clients to train a single shared model makes a good fit with our goal of cross-table pretraining. In this work, XTab leverages the distributed nature of federated learning to scale with a large number of pretraining tasks.

\vspace{-0.1in}
\section{Experiments} \vspace{-0.05in}
We evaluate the performance of XTab on supervised tabular learning tasks, including binary and multiclass classification and regression. We tested on the following pretraining settings:
\begin{itemize} \vspace{-0.1in}
\itemsep-0.3em 
\item XTab with various pretraining objectives, including reconstruction loss, contrastive loss, and supervised loss.
\item XTab with various transformer backbones, including FT-Transformer, Fastformer, and Saint-v. 
\item XTab with the transformer backbone partially- or fully-pretrained from other tasks.
\item XTab with different numbers of pretraining tasks.
\vspace{-0.1in}
\end{itemize}
During finetuning, we randomly initialize a new featurizer and projection head for each downstream task. All downstream tasks use the pretrained transformer backbone. We finetune all the model components using the training set of each downstream task. We included two different finetuning settings: 
\begin{itemize} \vspace{-0.1in}
\itemsep-0.3em 
\item Light finetuning: finetune XTab %\xingjian{Mention that the baseline models will also be fintuned for 3 epochs} 
for a fixed number of epochs (3 epochs). 
\item Heavy finetuning: finetune XTab with an early stopping patience of 3 epochs. The maximum number of epochs is set to infinity in this case. 
% \item Reduce the downstream training set by a certain ratio and finetune the pretrained Transformer on the reduced dataset.
\vspace{-0.1in}
\end{itemize}
For all finetuning settings, we retrieve the best model checkpoint based on validation scores, and use it to report the performance on the test data. The baseline models share the same model architecture and finetuning configurations as XTab, but with randomly initialized parameters instead of using the pretrained backbones. We find that XTab generally outperforms the baseline models in all scenarios and beats other deep learning models on tabular tasks.  Ablation study on the number of pretraining datasets is in Appendix \ref{app_num_tasks}. %\xingjian{Try to summarize the key findings}

\vspace{-0.1in}
\subsection{Datasets} \vspace{-0.05in}
We use the public OpenML-AutoML Benchmark (AMLB: \url{openml.github.io/automlbenchmark/}) \citep{gijsbers2022amlb} for pretraining and evaluation. AMLB is a recently proposed benchmark for automated machine learning, consisting of 104 tabular tasks (71 classification and 33 regression). We included the details of each dataset in Table~\ref{ds_stat} in the Appendix. Out of the 104 tabular datasets, we used 52 datasets for pretraining and the remaining 52 tasks for finetuning and evaluation. We split the pretraining and finetuning datasets by the alphabetical order of the task names (Table~\ref{ds_stat} in the Appendix).  % In AMLB, similar tasks may have similar names (e.g., ``QSAR-TID-10980" vs. ``QSAR-TID-11" \footnote{OpenML task id ``QSAR-TID-10980": 360933; ``QSAR-TID-11": 360932.}). This split ensures that the pretraining and finetuning datasets share minimal task similarity, and XTab learns general knowledge of tabular tasks.

\textbf{Data split:} For all downstream (or finetuning) tasks, AMLB reserves 10\% of the tabular data for testing. Over the remaining data, we randomly partition 87.5\% (7/8) into the training set and use 12.5\% (1/8) for validation. %we randomly partition the tabular data into 78.75\% training set, 11.25\% validation set, and 10\% test set. 
We repeated 5 trials with different test folds for all tabular datasets. All methods use the same split within the same trial. 

\textbf{Data pre-processing:} Following \citet{bahri2021scarf, somepalli2021saint, wang2022transtab}, we limit the discussion to tables with numerical and categorical columns. Each Category is represented by a distinct integer to index the embedding in the lookup table of the categorical featurizer (see Section \ref{Featurizers} for details). We normalized the numerical features by subtracting the mean and dividing them by the standard deviation. For regression tasks, we also apply the Standardization to the labels. The normalization parameters are calculated using the training set only to avoid information leakage. Missing entries are filled with the mean values of numerical columns, or treated as an additional category for categorical columns. 

\textbf{Table corruption:} Self-supervised learning objectives, including both contrastive and reconstruction losses, require a corrupted view of the input sample. In this work, we follow \citet{bahri2021scarf, rubachev2022revisiting} to randomly resample features and construct a corrupted sample. Specifically, we randomly select a fraction of features at each row of the table. Those features are corrupted by resampling from the empirical marginal distribution of the column. For all datasets, the corruption ratio was set to 60\% as suggested in \citet{bahri2021scarf}. In other words, for each sample $x$ and its corrupted view $\tilde{x}$, 60\% of entries are resampled whereas 40\% of features remain unchanged. 

\begin{figure}[ht]
\vskip 0in
\begin{center}
\centerline{\includegraphics[width=\columnwidth]{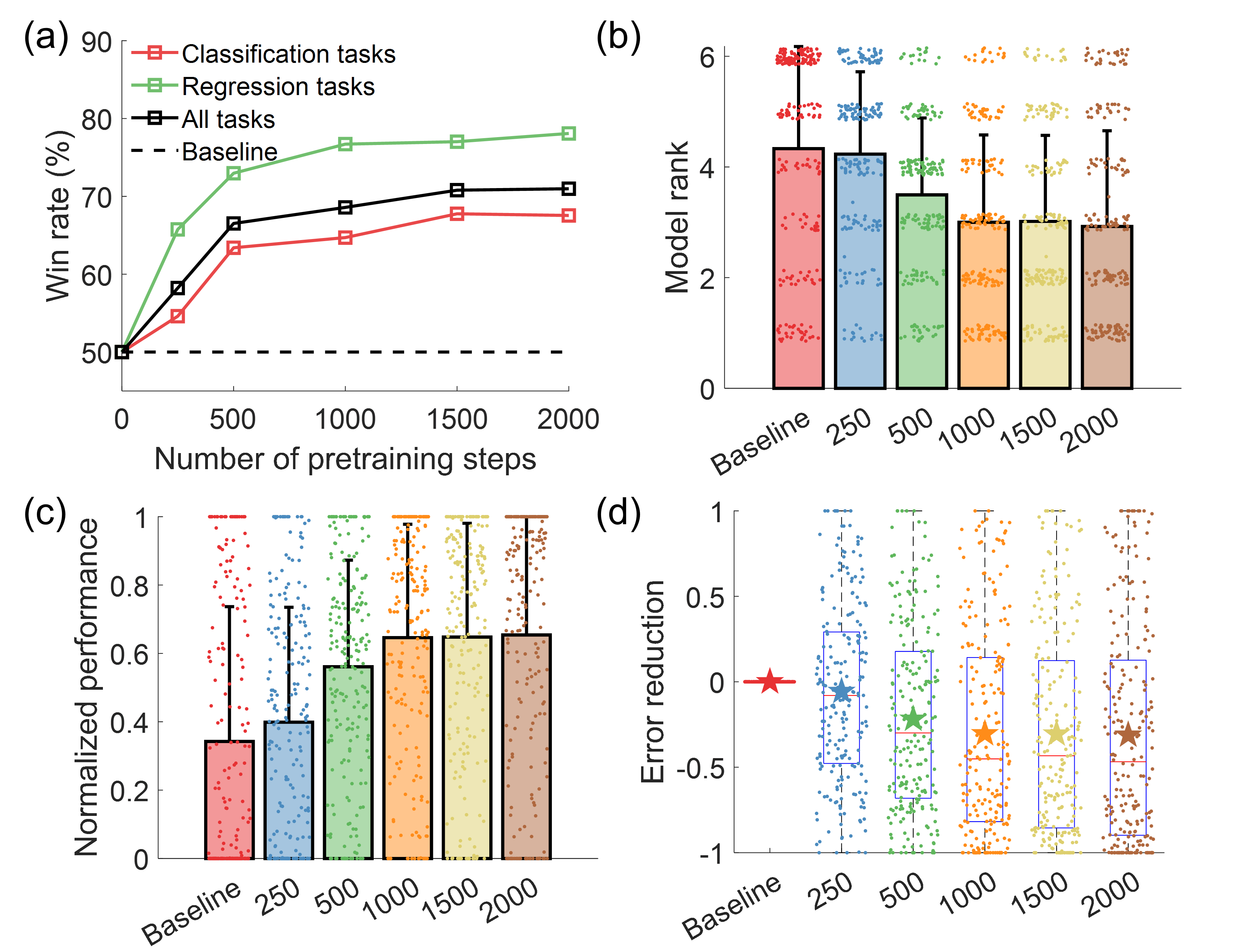}}
\vspace{-0.1in}
\caption{Tabular prediction performance of XTab using various evaluation criteria under the light finetuning setting. \textbf{(a)} The win rate of the pretrained transformer with respect to baseline. \textbf{(b)} The average rank of the models. \textbf{(c)} The normalized prediction performance. \textbf{(d)} The average error reduction rate compared to baseline.  Each dot indicates a trial of the downstream task (5 trials per dataset). The error bars show standard deviations in \textbf{(b)} and \textbf{(c)}. As the backbone is pretrained for more steps, we observe an increase in all evaluation criteria.}
\label{performance}
\end{center}
\vskip -0.5in
\end{figure}

\begin{figure}[ht]
\vskip 0in
\begin{center}
\centerline{\includegraphics[width=\columnwidth]{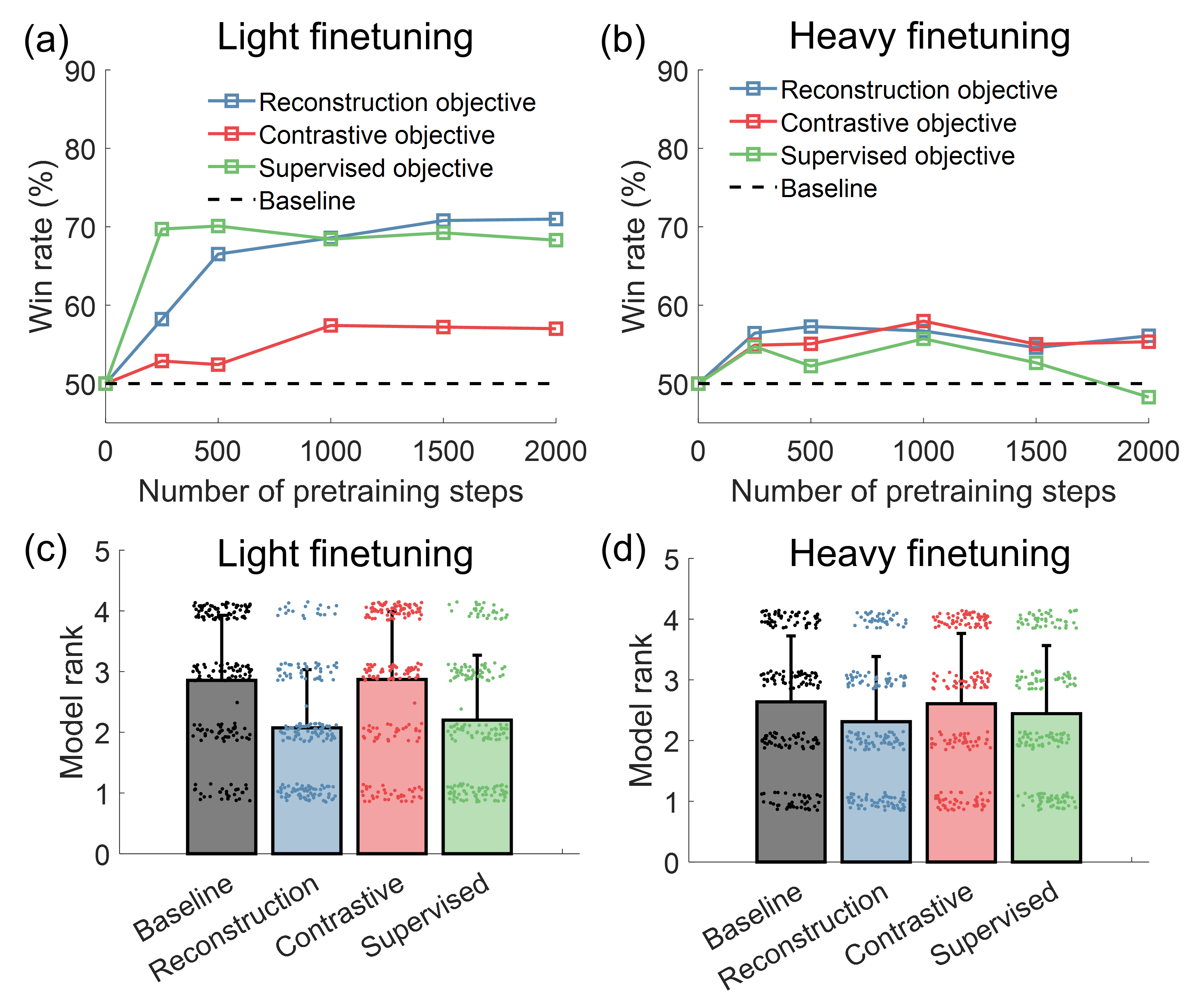}} 
\vspace{-0.1in}
\caption{Comparison of different pretraining objectives under the light \textbf{(a, c)} and heavy \textbf{(b, d)} finetuning settings. We show the win rate of XTab with different objectives with \textbf{(a)} light and \textbf{(b)} heavy  finetuning settings. We also compared the performance of pretraining objectives in terms of the model rank with \textbf{(c)} light and \textbf{(d)} heavy finetuning. We observe a consistent improvement of XTab compared to baseline models with all objectives. The reconstruction pretraining objective achieves the best performance, with 71.0\% win rate under light finetuning and 56.1\% for heavy finetuning at 2000 pretraining steps.}
\label{objectives}
\end{center}
\vskip -0.4in
\end{figure}

\vspace{-0.1in}
\subsection{Experimental setup} \vspace{-0.05in}
We used a federated pretraining setting as detailed in Section \ref{federated_pretrain}. Both pretraining and finetuning were performed on a cloud cluster of NVIDIA T4 GPUs (16 GB memory). We used about 30 thousand GPU hours for all experiments.
%\footnote{The code and sample pretrained checkpoints are attached to \url{anonymous.4open.science/r/XTab-F6CD}.}

\textbf{Model configuration and training:} 
Our default model configuration of transformer variants is the same as \citet{gorishniy2021revisiting}, with 3 transformer blocks, a feature embedding size of 192 and 8 attention heads. The feed forward networks (Figure \ref{model_structure}) have two layers with the same size as the embedding. We apply a dropout ratio of 20\% to attention layers and 10\% for feed forward networks. We use ReGLU \citep{shazeer2020glu} as the activation function and layer normalization \citep{ba2016layer} in the feed forward layers. The projection heads are ReLU networks with 2 layers and a hidden dimension of 192. All model components use \textit{Kaiming} initialization \citep{he2015delving} with the bias terms fixed at zeros. 

The batch size is fixed at 128 for both pretraining and finetuning. Both stages use AdamW as the optimizer, with a learning rate of 1e-4. Following \citet{gorishniy2021revisiting, rubachev2022revisiting}, we also apply a weight decay of 1e-5 to all components excluding featurizers, $\textsf{[CLS]}$ tokens, layer normalization and bias terms.
 
\textbf{Evaluation metrics:}
We choose the evaluation metrics as suggested by AMLB \citep{gijsbers2022amlb}. We use  root mean-squared error (RMSE) for regression tasks, area under the receiver operating characteristic curve (AUC) for binary classification, and log loss for multi-class classification. The same evaluation metrics are applied to validation sets for early stopping. The efficacy of the pretrained transformer backbones is estimated by the downstream performance.

% \textbf{Finetuning time:}
% Throughout all experiments, finetuning from a pretrained backbone roughly uses the same wall-clock time as training from scratch ($<$3\% difference). 
\vspace{-0.1in}
\subsection{Comparison with baseline transformers} \vspace{-0.05in}

\textbf{Cross-table pretraining improves downstream task performance.} As shown in Figure \ref{performance}, we compare the downstream prediction performance of FT-Transformer before (baseline) and after cross-table pretraining. Reconstruction objective is used for pretraining and all downstream tasks are finetuned for 3 epochs (light finetuning). We checkpoint the pretrained backbone after a certain number of pretraining steps and finetune downstream tasks from various checkpoints (250/500/1000/1500/2000). In Figure \ref{performance}(a), we show the win rate of the pretrained transformer on all downstream tasks with respect to baseline. Both classification and regression tasks benefit from our proposed cross-table pretraining. As the backbone is pretrained for more steps, we observe an increase in the win rate. We also calculate the rank of the model for each downstream task (Figure \ref{performance}(b)). Model rank is an integer from 1 to 6, with a lower number indicating better performance. Equal values are assigned a rank that is the average of the ranks of those values. The rank of the model improves with XTab pretraining. To further validate the advantage of XTab over transformers without cross-table pretraining, we further look into the normalized prediction performance and error reduction rate (Figure \ref{performance}(c, d)). We min-max normalize the prediction performance of all models, such that the worst model receives a score of 0 and the best model receives 1. Similarly, errors are also normalized to the best and worst models. Negative numbers indicate a model with lower error ($1 - \text{AUC scores}$ for binary classification) or loss (log loss for multiclass classification and RMSE for regression) than baseline. The mean error (or loss) is indicated by the stars. FT-Transformers pretrained with XTab on average obtain higher normalized performance and reduced error compared to traditional random initialization.

\begin{figure}[ht]
\vskip 0in
\begin{center}
\centerline{\includegraphics[width=\columnwidth]{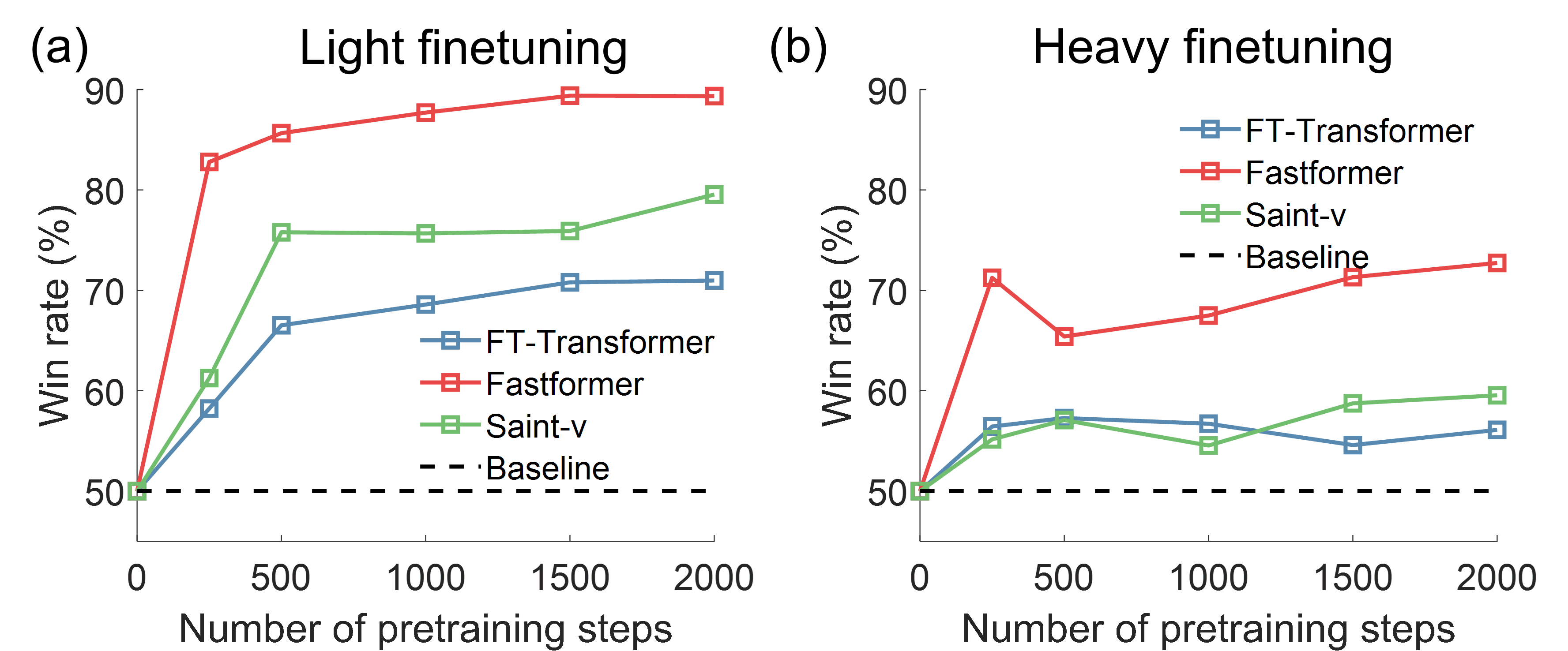}}
\vspace{-0.1in}
\caption{XTab with transformer variants including FT-Transformer, Fastformer, and Saint-v. We use different transformer models as the shared backbone in XTab. We calculate the win rate of the pretrained backbone over randomly initialized transformers. \textbf{(a)} shows the results for light finetuning and \textbf{(b)} represents heavy finetuning. FT-Transformer, Fastformer, and Saint-v all benefit from our proposed cross-table pretraining, achieving $>$50\% win rate in all experiments.}
\label{backbone}
\end{center}
\vskip -0.3in
\vspace{-0.2in}
\end{figure}

\textbf{XTab with different pretraining objectives and finetuning settings.} We extensively test XTab with various pretraining objectives and finetuning settings. Figure \ref{objectives} summarizes the downstream performance using reconstruction, contrastive and supervised objectives as described in Section \ref{obj}. We use FT-Transformer as the backbone. Figure \ref{objectives}(a, b) plot the win rate of XTab under the light and heavy finetuning settings, respectively. We finetune on all downstream tasks for 3 epochs with light finetuning, and use an early stopping patience of 3 for heavy finetuning. We observe a consistent improvement of XTab over the baseline with no cross-table pretraining. The advantage of XTab is more significant in the light finetuning setting compared to heavy finetuning. For example, XTab with the reconstruction objective achieves a 71.0\% win rate with light finetuning, but only 56.1\% with heavy finetuning. The difference is caused by catastrophic forgetting of deep models \citep{ramasesh2021effect, kaushik2021understanding}. As tabular transformers are relatively small ($<$1M parameters for the FT-Transformer backbone), they are more vulnerable to catastrophic forgetting during the finetuning phase. It is possible to alleviate this issue with additional techniques \citep{ramasesh2021effect, kaushik2021understanding}, but this is outside the scope of the paper. Figure \ref{objectives}(c, d) compare different objectives by ranking the models with light and heavy finetuning. All approaches are pretrained for 2000 steps. Each dot in Figure \ref{objectives}(c, d) represents a trial of downstream experiments (5 trials per dataset) and error bars indicate the standard deviations across trials. The advantage of cross-table pretraining is shown by a win rate $>$50\% and a model rank value lower than the baseline. A more detailed comparison involving the normalized performance and error reduction rate is presented in Appendix \ref{app_supp}. We conclude that XTab consistently enhances the downstream performance of tabular transformers across multiple pretraining objectives and finetuning settings. Among all pretraining objectives tested, reconstruction loss performs better than contrastive or supervised losses. 

% \begin{figure}[h]
% \vskip 0in
% \begin{center}
% \centerline{\includegraphics[width=\columnwidth]{figures/shareable_blocks.png}}
% \caption{Comparison of XTab with various pretrained components in FT-Transformer. We run this study to understand which component carries general knowledge of tabular tasks and benefits from cross-table pretraining. Several settings are tested, sharing the first block of Transformer, all blocks, $\textsf{[CLS]}$ token, all blocks with $\textsf{[CLS]}$ token, or no component (baseline). Performance is compared in terms of \textbf{(a)} win rate and \textbf{(b)} model rank with light finetuning. Pretraining on the Transformer blocks leads to improved performance, whereas sharing the task-specific $\textsf{[CLS]}$ token is hardly beneficial.}
% \label{shareable_blocks}
% \end{center}
% %\vskip -0.4in
% \end{figure}

\textbf{XTab is applicable to various types of transformers.} %\xingjian{Consider to not use ``agnostic"}
XTab offers a framework to pretrain the shared model components across tabular tasks. Therefore, the choice of transformer backbone is flexible, as long as the model can process tables with variable columns. In Figure \ref{backbone}, we plug three transformer variants into XTab including FT-Transformer, Fastformer, and Saint-v. The explanation of transformer backbones can be found in Section \ref{method_backbone}. We pretrain all transformers using reconstruction objective, and finetune on the downstream tasks with the light and heavy settings, Figure \ref{backbone}(a, b). We show that XTab is applicable to various types of transformers and all models benefit from the proposed cross-table pretraining, achieving a higher win rate compared to the baseline.

% \textbf{Identifying the shareable components in XTab.} In XTab, we separate a model into task-specific components (e.g., featurizers and projection heads) and shareable components (Transformer blocks). Only the shareable components are pretrained and contain general knowledge of tabular learning. Therefore, identifying the shareable (or pretrainable) components is critical to the success of cross-table pretraining. In Figure \ref{shareable_blocks}, we run an experiment to pretrain on different FT-Transformer components with the supervised objective. For example, pretraining tasks may share only the first Transformer block and the later two blocks are marked as task-specific. We also let the pretraining tasks share all Transformer blocks, $\textsf{[CLS]}$ token, and all blocks with $\textsf{[CLS]}$ token. As expected, pretraining on the $\textsf{[CLS]}$ token does not lead to improved downstream performance, since $\textsf{[CLS]}$ token is directly related to downstream prediction and thereby highly task-specific. From Figure \ref{shareable_blocks}, we find that it is most beneficial to pretrain on all Transformer blocks without the $\textsf{[CLS]}$ token. Featurizers and projection heads are not shareable since the input/output spaces can be different across tasks.

Additional experimental results are presented in the Appendix. In Appendix \ref{appendix_sharable_component}, we pretrain on different components of transformers to identify the shareable components in XTab. In Appendix \ref{app_subsample}, we look into the downstream performance with only a portion of the training set used for finetuning. In Appendix \ref{app_num_tasks}, we compare XTab backbone pretrained on different numbers of tasks and find that more pretraining tasks lead to improved performance. In Appendix \ref{app_fedavg}, we study the federated pretraining setting by changing the number of local updates per global aggregation (i.e., $N$), and find that larger $N$ leads to reduced downstream performance.

\vspace{-0.1in}
\subsection{Performance compared to traditional baselines} \vspace{-0.05in}

\begin{table}[ht]
\caption{Comparison of tabular prediction performance with default model configuration and hyperparameter optimization (HPO). Mean training time and model rank ($\pm$ standard deviation) are calculated across 84 datasets from AutoML Benchmark. We perform 5 independent trials for each task. XTab outperforms its counterpart FTT in all scenarios thanks to cross-table pretraining, whereas CatBoost is the overall best model. The best overall method (CatBoost) and the best deep learning approach (XTab-best) are highlighted in \textbf{bold}. }
\label{t1}
% \vskip 0.15in
\begin{center}
\begin{tabular}{llccr}
\toprule
& Methods & Time (s) & Rank  \\
\midrule
\multirow{13}{*}{ \rotatebox{90}{Default hyperparameter} }
& RF & 66.8$^\dagger$  & 7.14 $\pm$ 3.81 \\
& XGBoost & 43.1$^\dagger$ & 5.06 $\pm$ 3.08\\
& LightGBM & 23.9$^\dagger$ & 5.23 $\pm$ 3.25 \\
& \textbf{CatBoost} & \textbf{322.8}$^\dagger$ & \textbf{2.98 $\pm$ 2.66} \\
\cmidrule{2-4}
& FastAI & 89.6 & 7.24 $\pm$ 3.44 \\
& NN & 188.8 &  7.40 $\pm$ 3.43\\
\cmidrule{2-4}
& TransTab-sl$^*$ & 539.7 &  11.04 $\pm$ 2.75\\
& TransTab-cl$^*$ & 312.0 &  10.79 $\pm$ 3.00\\
\cmidrule{2-4}
& FTT-l & 189.2 & 10.19 $\pm$ 2.43 \\
& XTab-l & 189.8 & 9.21 $\pm$ 2.57 \\
\cmidrule{2-4}
& FTT-h & 532.5 & 7.29 $\pm$ 2.20 \\
& XTab-h & 506.3 & 6.93 $\pm$ 2.09 \\
\cmidrule{2-4}
& FTT-best & 810.9 &  4.94 $\pm$ 2.25 \\
& \textbf{XTab-best} & \textbf{755.9} &  \textbf{4.39 $\pm$ 2.36} \\
\bottomrule
\toprule
%\specialrule{.2em}{.1em}{.1em} 
\multirow{8}{*}{ \rotatebox{90}{HPO} }
& RF & 1084.4$^\dagger$  & 5.00 $\pm$ 2.40 \\
& XGBoost & 862.3$^\dagger$ & 3.69 $\pm$ 2.45\\
& LightGBM & 285.0$^\dagger$ & 4.40 $\pm$ 1.93 \\
& \textbf{CatBoost} & \textbf{1529.3}$^\dagger$ & \textbf{3.25 $\pm$ 2.10} \\
\cmidrule{2-4}
& FastAI & 549.7 & 5.24 $\pm$ 2.38 \\
& NN & 1163.5 &  5.32 $\pm$ 2.20\\
\cmidrule{2-4}
& FTT & 2221.1 &  4.58 $\pm$ 2.08\\
& \textbf{XTab} & \textbf{2335.3} &  \textbf{4.51 $\pm$ 2.00} \\
\bottomrule
\multicolumn{4}{l}{\footnotesize{$^\dagger$ CPU training time.}} \\
\multicolumn{4}{l}{\footnotesize{$^*$ Only evaluated on classification tasks.}} \\
\end{tabular}
\end{center}
\vskip -0.3in
\vspace{-0.1in}
\end{table}

To compare the performance of XTab and various tabular models, we run experiments on the full AutoML Benchmark \citep{gijsbers2022amlb}. We split the benchmark into 2 folds, each consisting of 52 tabular datasets. We pretrain on fold \#1 and evaluate the downstream performance on fold \#2 and vice versa. We pretrain XTab with the FT-Transformer backbone using reconstruction loss. 20 datasets are excluded since they could not fit into the GPU memory (16 GB, see Table \ref{ds_stat} in the Appendix for details). We report the performance on the remaining 84 tasks. In addition to XTab, we include the following methods:

\textbf{Tree-based models:} Tree-based models provide strong performance on tabular tasks \citep{grinsztajn2022tree}. We include Random Forest (RF) and gradient-boosted tree variants: XGBoost \citep{chen2016xgboost}, LightGBM \citep{ke2017lightgbm} and CatBoost \citep{dorogush2018catboost}. \textbf{Neural networks:} We include the AutoGluon neural networks implemented on top of PyTorch \citep{erickson2020autogluon} and the FastAI tabular model \citep{howard2020fastai}. \textbf{Transformers:} We include the FT-Transformer which is a direct counterpart of XTab without pretraining. The finetuning settings of FTT/XTab include light (FTT-l/XTab-l) and heavy (FTT-h/XTab-h) finetuning as described above. We further introduce FTT-best/XTab-best, which incorporates an early-stopping patience of 20 and model soup of the top 3 checkpoints \citep{wortsman2022model} to achieve better performance. TransTab is included for comparison on classification tasks (regression not enabled yet with TransTab) under the supervised learning (TransTab-sl) and contrastive learning (TransTab-cl) settings \citep{wang2022transtab}. Please refer to Appendix \ref{app_transtab} for how the TransTab ranks are calculated, and Table \ref{t1_class} for results on classification tasks only. %Detailed model configurations are available in Appendix \ref{benchmark_configurations}.

Table \ref{t1} shows the performance of models with the default hyperparameters and hyperparameter optimization (HPO). With the default hyperparameter, we pretrain XTab for 2000 rounds, whereas the number of pretraining rounds is tuned under the HPO setting. We use the AutoGluon default hyperparameters for tree-based models as they outperform the official defaults to give a strong baseline \citep{erickson2020autogluon}. CatBoost is the state-of-the-art model on tabular tasks, which agrees with the recent finding in \citet{grinsztajn2022tree}. With cross-table pretraining, XTab improves the performance over FTT under light (FTT-l/XTab-l) and heavy (FTT-h/XTab-h) finetuning. Using more finetuning time, XTab-best achieves second place in the benchmark and beats other deep learning models. The success of XTab using the default configuration ensures that the pretrained backbone is widely applicable to tabular tasks, without the need for case-by-case tuning.

With HPO, we randomly search for data-specific hyperparameters on the validation performance. The detailed search space of each model is in Appendix \ref{benchmark_configurations}. We allow a maximum number of 100 HPO trials within a 1-hour time budget. Table \ref{t1} shows that gradient-boosted trees (i.e., XGBoost, LightGBM, CatBoost) achieve higher ranking with HPO, since they are generally faster to train. The search space is also smaller for tree models as they have fewer meaningful hyperparameters and well-known highly performant search spaces. The ranks are calculated separately for default hyperparameters and HPO and  are not comparable across the two settings. The advantage of XTab over FTT increases as we allocate less training time for downstream tasks (XTab-l $\leftarrow$ XTab-h $\leftarrow$ XTab-best $\leftarrow$ XTab with HPO). Therefore, one should use pretrained foundation models instead of randomly initialized weights for tabular transformers, especially with a tight training budget. 

\vspace{-0.1in}
\section{Conclusion} \vspace{-0.05in}
In this paper, we present XTab to improve the performance of deep tabular models. XTab pretrains tabular transformers with a diverse collection of data tables, and can improve the tabular prediction performance of an unseen table from arbitrary domains. XTab handles the cross-table variations by separating the models into data-specific and shared components, and encourages the shared components to learn general knowledge for tabular prediction. We also propose to combine self-supervised pretraining with federated learning to improve pretraining efficiency, where client-side nodes perform table reconstruction tasks followed by backbone averaging updates at the server. Our results suggest that finetuning from the pretrained transformer is superior to training tabular transformers from scratch. One limitation of XTab is that it still falls behind CatBoost. This motivates  future works on bridging the gap between pretrained tabular deep learning models and tree models. Another interesting  direction is to combine XTab with language/vision foundation models for improving multimodal learning.

\vspace{-0.1in}
\section*{Software and Data} \vspace{-0.05in}
The AutoML Benchmark (AMLB) is publicly available at \url{openml.github.io/automlbenchmark}. The code and sample pretrained checkpoints are attached to \url{https://github.com/BingzhaoZhu/XTab}.

% \section*{Acknowledgements}

% \textbf{Blocked for blind review.}


\bibliography{example_paper}

\begin{thebibliography}{38}
\providecommand{\natexlab}[1]{#1}
\providecommand{\url}[1]{\texttt{#1}}
\expandafter\ifx\csname urlstyle\endcsname\relax
  \providecommand{\doi}[1]{doi: #1}\else
  \providecommand{\doi}{doi: \begingroup \urlstyle{rm}\Url}\fi

\bibitem[Aghajanyan et~al.(2021)Aghajanyan, Gupta, Shrivastava, Chen,
  Zettlemoyer, and Gupta]{aghajanyan2021muppet}
Aghajanyan, A., Gupta, A., Shrivastava, A., Chen, X., Zettlemoyer, L., and
  Gupta, S.
\newblock Muppet: Massive multi-task representations with pre-finetuning.
\newblock \emph{arXiv preprint arXiv:2101.11038}, 2021.

\bibitem[Ba et~al.(2016)Ba, Kiros, and Hinton]{ba2016layer}
Ba, J.~L., Kiros, J.~R., and Hinton, G.~E.
\newblock Layer normalization.
\newblock \emph{arXiv preprint arXiv:1607.06450}, 2016.

\bibitem[Bahri et~al.(2021)Bahri, Jiang, Tay, and Metzler]{bahri2021scarf}
Bahri, D., Jiang, H., Tay, Y., and Metzler, D.
\newblock Scarf: Self-supervised contrastive learning using random feature
  corruption.
\newblock \emph{arXiv preprint arXiv:2106.15147}, 2021.

\bibitem[Bommasani et~al.(2021)Bommasani, Hudson, Adeli, Altman, Arora, von
  Arx, Bernstein, Bohg, Bosselut, Brunskill,
  et~al.]{bommasani2021opportunities}
Bommasani, R., Hudson, D.~A., Adeli, E., Altman, R., Arora, S., von Arx, S.,
  Bernstein, M.~S., Bohg, J., Bosselut, A., Brunskill, E., et~al.
\newblock On the opportunities and risks of foundation models.
\newblock \emph{arXiv preprint arXiv:2108.07258}, 2021.

\bibitem[Chen \& Guestrin(2016)Chen and Guestrin]{chen2016xgboost}
Chen, T. and Guestrin, C.
\newblock Xgboost: A scalable tree boosting system.
\newblock In \emph{Proceedings of the 22nd acm sigkdd international conference
  on knowledge discovery and data mining}, pp.\  785--794, 2016.

\bibitem[Chen et~al.(2020)Chen, Kornblith, Norouzi, and Hinton]{chen2020simple}
Chen, T., Kornblith, S., Norouzi, M., and Hinton, G.
\newblock A simple framework for contrastive learning of visual
  representations.
\newblock In \emph{International conference on machine learning}, pp.\
  1597--1607. PMLR, 2020.

\bibitem[Collins et~al.(2022)Collins, Hassani, Mokhtari, and
  Shakkottai]{collins2022fedavg}
Collins, L., Hassani, H., Mokhtari, A., and Shakkottai, S.
\newblock Fedavg with fine tuning: Local updates lead to representation
  learning.
\newblock \emph{arXiv preprint arXiv:2205.13692}, 2022.

\bibitem[Devlin et~al.(2018)Devlin, Chang, Lee, and Toutanova]{devlin2018bert}
Devlin, J., Chang, M.-W., Lee, K., and Toutanova, K.
\newblock Bert: Pre-training of deep bidirectional transformers for language
  understanding.
\newblock \emph{arXiv preprint arXiv:1810.04805}, 2018.

\bibitem[Dorogush et~al.(2018)Dorogush, Ershov, and
  Gulin]{dorogush2018catboost}
Dorogush, A.~V., Ershov, V., and Gulin, A.
\newblock Catboost: gradient boosting with categorical features support.
\newblock \emph{arXiv preprint arXiv:1810.11363}, 2018.

\bibitem[Dosovitskiy et~al.(2020)Dosovitskiy, Beyer, Kolesnikov, Weissenborn,
  Zhai, Unterthiner, Dehghani, Minderer, Heigold, Gelly,
  et~al.]{dosovitskiy2020image}
Dosovitskiy, A., Beyer, L., Kolesnikov, A., Weissenborn, D., Zhai, X.,
  Unterthiner, T., Dehghani, M., Minderer, M., Heigold, G., Gelly, S., et~al.
\newblock An image is worth 16x16 words: Transformers for image recognition at
  scale.
\newblock \emph{arXiv preprint arXiv:2010.11929}, 2020.

\bibitem[Erickson et~al.(2020)Erickson, Mueller, Shirkov, Zhang, Larroy, Li,
  and Smola]{erickson2020autogluon}
Erickson, N., Mueller, J., Shirkov, A., Zhang, H., Larroy, P., Li, M., and
  Smola, A.
\newblock Autogluon-tabular: Robust and accurate automl for structured data.
\newblock \emph{arXiv preprint arXiv:2003.06505}, 2020.

\bibitem[Gijsbers et~al.(2022)Gijsbers, Bueno, Coors, LeDell, Poirier, Thomas,
  Bischl, and Vanschoren]{gijsbers2022amlb}
Gijsbers, P., Bueno, M.~L., Coors, S., LeDell, E., Poirier, S., Thomas, J.,
  Bischl, B., and Vanschoren, J.
\newblock Amlb: an automl benchmark.
\newblock \emph{arXiv preprint arXiv:2207.12560}, 2022.

\bibitem[Gorishniy et~al.(2021)Gorishniy, Rubachev, Khrulkov, and
  Babenko]{gorishniy2021revisiting}
Gorishniy, Y., Rubachev, I., Khrulkov, V., and Babenko, A.
\newblock Revisiting deep learning models for tabular data.
\newblock \emph{Advances in Neural Information Processing Systems},
  34:\penalty0 18932--18943, 2021.

\bibitem[Grinsztajn et~al.(2022)Grinsztajn, Oyallon, and
  Varoquaux]{grinsztajn2022tree}
Grinsztajn, L., Oyallon, E., and Varoquaux, G.
\newblock Why do tree-based models still outperform deep learning on typical
  tabular data?
\newblock In \emph{Thirty-sixth Conference on Neural Information Processing
  Systems Datasets and Benchmarks Track}, 2022.

\bibitem[He et~al.(2015)He, Zhang, Ren, and Sun]{he2015delving}
He, K., Zhang, X., Ren, S., and Sun, J.
\newblock Delving deep into rectifiers: Surpassing human-level performance on
  imagenet classification.
\newblock In \emph{Proceedings of the IEEE international conference on computer
  vision}, pp.\  1026--1034, 2015.

\bibitem[He et~al.(2022)He, Chen, Xie, Li, Doll{\'a}r, and
  Girshick]{he2022masked}
He, K., Chen, X., Xie, S., Li, Y., Doll{\'a}r, P., and Girshick, R.
\newblock Masked autoencoders are scalable vision learners.
\newblock In \emph{Proceedings of the IEEE/CVF Conference on Computer Vision
  and Pattern Recognition}, pp.\  16000--16009, 2022.

\bibitem[Hollmann et~al.(2022)Hollmann, M{\"u}ller, Eggensperger, and
  Hutter]{hollmann2022tabpfn}
Hollmann, N., M{\"u}ller, S., Eggensperger, K., and Hutter, F.
\newblock Tabpfn: A transformer that solves small tabular classification
  problems in a second.
\newblock \emph{arXiv preprint arXiv:2207.01848}, 2022.

\bibitem[Howard \& Gugger(2020)Howard and Gugger]{howard2020fastai}
Howard, J. and Gugger, S.
\newblock Fastai: a layered api for deep learning.
\newblock \emph{Information}, 11\penalty0 (2):\penalty0 108, 2020.

\bibitem[Kaushik et~al.(2021)Kaushik, Gain, Kortylewski, and
  Yuille]{kaushik2021understanding}
Kaushik, P., Gain, A., Kortylewski, A., and Yuille, A.
\newblock Understanding catastrophic forgetting and remembering in continual
  learning with optimal relevance mapping.
\newblock \emph{arXiv preprint arXiv:2102.11343}, 2021.

\bibitem[Ke et~al.(2017)Ke, Meng, Finley, Wang, Chen, Ma, Ye, and
  Liu]{ke2017lightgbm}
Ke, G., Meng, Q., Finley, T., Wang, T., Chen, W., Ma, W., Ye, Q., and Liu,
  T.-Y.
\newblock Lightgbm: A highly efficient gradient boosting decision tree.
\newblock \emph{Advances in neural information processing systems}, 30, 2017.

\bibitem[Kurin et~al.(2022)Kurin, De~Palma, Kostrikov, Whiteson, and
  Kumar]{kurin2022defense}
Kurin, V., De~Palma, A., Kostrikov, I., Whiteson, S., and Kumar, M.~P.
\newblock In defense of the unitary scalarization for deep multi-task learning.
\newblock \emph{arXiv preprint arXiv:2201.04122}, 2022.

\bibitem[Levin et~al.(2022)Levin, Cherepanova, Schwarzschild, Bansal, Bruss,
  Goldstein, Wilson, and Goldblum]{levin2022transfer}
Levin, R., Cherepanova, V., Schwarzschild, A., Bansal, A., Bruss, C.~B.,
  Goldstein, T., Wilson, A.~G., and Goldblum, M.
\newblock Transfer learning with deep tabular models.
\newblock \emph{arXiv preprint arXiv:2206.15306}, 2022.

\bibitem[Li et~al.(2019)Li, Huang, Yang, Wang, and Zhang]{li2019convergence}
Li, X., Huang, K., Yang, W., Wang, S., and Zhang, Z.
\newblock On the convergence of fedavg on non-iid data.
\newblock \emph{arXiv preprint arXiv:1907.02189}, 2019.

\bibitem[Liu et~al.(2021)Liu, Lin, Cao, Hu, Wei, Zhang, Lin, and
  Guo]{liu2021swin}
Liu, Z., Lin, Y., Cao, Y., Hu, H., Wei, Y., Zhang, Z., Lin, S., and Guo, B.
\newblock Swin transformer: Hierarchical vision transformer using shifted
  windows.
\newblock In \emph{Proceedings of the IEEE/CVF International Conference on
  Computer Vision}, pp.\  10012--10022, 2021.

\bibitem[Majmundar et~al.(2022)Majmundar, Goyal, Netrapalli, and
  Jain]{majmundar2022met}
Majmundar, K., Goyal, S., Netrapalli, P., and Jain, P.
\newblock Met: Masked encoding for tabular data.
\newblock \emph{arXiv preprint arXiv:2206.08564}, 2022.

\bibitem[McMahan et~al.(2017)McMahan, Moore, Ramage, Hampson, and
  y~Arcas]{mcmahan2017communication}
McMahan, B., Moore, E., Ramage, D., Hampson, S., and y~Arcas, B.~A.
\newblock Communication-efficient learning of deep networks from decentralized
  data.
\newblock In \emph{Artificial intelligence and statistics}, pp.\  1273--1282.
  PMLR, 2017.

\bibitem[Ramasesh et~al.(2021)Ramasesh, Lewkowycz, and
  Dyer]{ramasesh2021effect}
Ramasesh, V.~V., Lewkowycz, A., and Dyer, E.
\newblock Effect of scale on catastrophic forgetting in neural networks.
\newblock In \emph{International Conference on Learning Representations}, 2021.

\bibitem[Rubachev et~al.(2022)Rubachev, Alekberov, Gorishniy, and
  Babenko]{rubachev2022revisiting}
Rubachev, I., Alekberov, A., Gorishniy, Y., and Babenko, A.
\newblock Revisiting pretraining objectives for tabular deep learning.
\newblock \emph{arXiv preprint arXiv:2207.03208}, 2022.

\bibitem[Shazeer(2020)]{shazeer2020glu}
Shazeer, N.
\newblock Glu variants improve transformer.
\newblock \emph{arXiv preprint arXiv:2002.05202}, 2020.

\bibitem[Somepalli et~al.(2021)Somepalli, Goldblum, Schwarzschild, Bruss, and
  Goldstein]{somepalli2021saint}
Somepalli, G., Goldblum, M., Schwarzschild, A., Bruss, C.~B., and Goldstein, T.
\newblock Saint: Improved neural networks for tabular data via row attention
  and contrastive pre-training.
\newblock \emph{arXiv preprint arXiv:2106.01342}, 2021.

\bibitem[Ucar et~al.(2021)Ucar, Hajiramezanali, and Edwards]{ucar2021subtab}
Ucar, T., Hajiramezanali, E., and Edwards, L.
\newblock Subtab: Subsetting features of tabular data for self-supervised
  representation learning.
\newblock \emph{Advances in Neural Information Processing Systems},
  34:\penalty0 18853--18865, 2021.

\bibitem[Vaswani et~al.(2017)Vaswani, Shazeer, Parmar, Uszkoreit, Jones, Gomez,
  Kaiser, and Polosukhin]{vaswani2017attention}
Vaswani, A., Shazeer, N., Parmar, N., Uszkoreit, J., Jones, L., Gomez, A.~N.,
  Kaiser, {\L}., and Polosukhin, I.
\newblock Attention is all you need.
\newblock \emph{Advances in neural information processing systems}, 30, 2017.

\bibitem[Wang \& Sun(2022)Wang and Sun]{wang2022transtab}
Wang, Z. and Sun, J.
\newblock Transtab: Learning transferable tabular transformers across tables.
\newblock \emph{arXiv preprint arXiv:2205.09328}, 2022.

\bibitem[Winkelmolen et~al.(2020)Winkelmolen, Ivkin, Bozkurt, and
  Karnin]{winkelmolen2020practical}
Winkelmolen, F., Ivkin, N., Bozkurt, H.~F., and Karnin, Z.
\newblock Practical and sample efficient zero-shot hpo.
\newblock \emph{arXiv preprint arXiv:2007.13382}, 2020.

\bibitem[Wortsman et~al.(2022)Wortsman, Ilharco, Gadre, Roelofs, Gontijo-Lopes,
  Morcos, Namkoong, Farhadi, Carmon, Kornblith, et~al.]{wortsman2022model}
Wortsman, M., Ilharco, G., Gadre, S.~Y., Roelofs, R., Gontijo-Lopes, R.,
  Morcos, A.~S., Namkoong, H., Farhadi, A., Carmon, Y., Kornblith, S., et~al.
\newblock Model soups: averaging weights of multiple fine-tuned models improves
  accuracy without increasing inference time.
\newblock In \emph{International Conference on Machine Learning}, pp.\
  23965--23998. PMLR, 2022.

\bibitem[Wu et~al.(2021)Wu, Wu, Qi, Huang, and Xie]{wu2021fastformer}
Wu, C., Wu, F., Qi, T., Huang, Y., and Xie, X.
\newblock Fastformer: Additive attention can be all you need.
\newblock \emph{arXiv preprint arXiv:2108.09084}, 2021.

\bibitem[Yin et~al.(2020)Yin, Neubig, Yih, and Riedel]{yin2020tabert}
Yin, P., Neubig, G., Yih, W.-t., and Riedel, S.
\newblock Ta{BERT}: Pretraining for joint understanding of textual and tabular
  data.
\newblock \emph{arXiv preprint arXiv:2005.08314}, 2020.

\bibitem[Yoon et~al.(2020)Yoon, Zhang, Jordon, and van~der
  Schaar]{yoon2020vime}
Yoon, J., Zhang, Y., Jordon, J., and van~der Schaar, M.
\newblock Vime: Extending the success of self-and semi-supervised learning to
  tabular domain.
\newblock \emph{Advances in Neural Information Processing Systems},
  33:\penalty0 11033--11043, 2020.

\end{thebibliography}
\bibliographystyle{icml2023}

%%%%%%%%%%%%%%%%%%%%%%%%%%%%%%%%%%%%%%%%%%%%%%%%%%%%%%%%%%%%%%%%%%%%%%%%%%%%%%%
%%%%%%%%%%%%%%%%%%%%%%%%%%%%%%%%%%%%%%%%%%%%%%%%%%%%%%%%%%%%%%%%%%%%%%%%%%%%%%%
% APPENDIX
%%%%%%%%%%%%%%%%%%%%%%%%%%%%%%%%%%%%%%%%%%%%%%%%%%%%%%%%%%%%%%%%%%%%%%%%%%%%%%%
%%%%%%%%%%%%%%%%%%%%%%%%%%%%%%%%%%%%%%%%%%%%%%%%%%%%%%%%%%%%%%%%%%%%%%%%%%%%%%%
\newpage
\appendix
\onecolumn
\section{XTab performance with various pretraining/finetuning settings} \label{app_supp}
\begin{figure}[H]
\vskip 0.2in
\begin{center}
\centerline{\includegraphics[width=16cm]{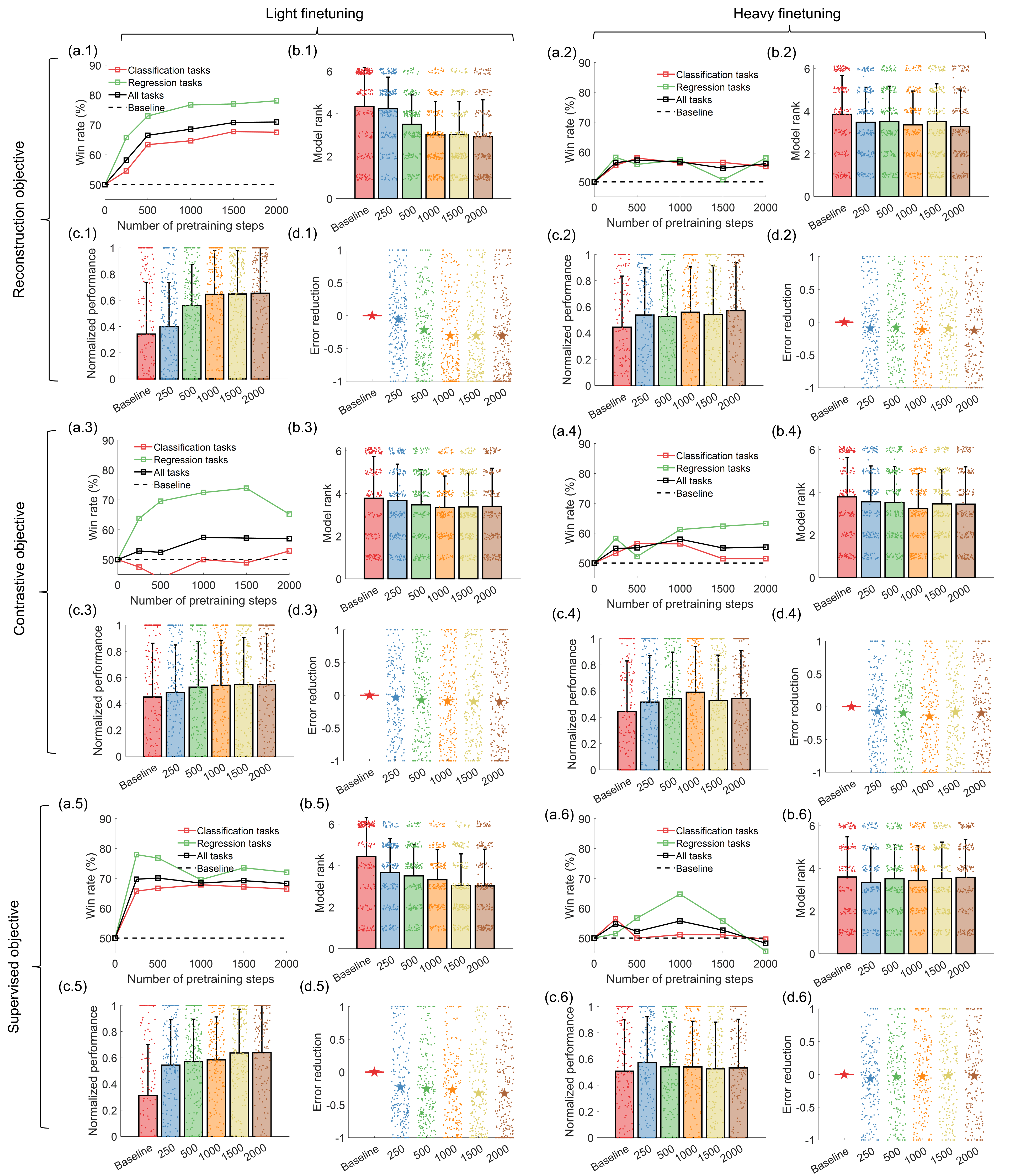}}
\caption{The figure is similar to Figure \ref{performance} in the main paper, but contains more pretraining/finetuning configurations. See the caption and explanation there for more details.}
\label{performance_supp}
\end{center}
\vskip -0.2in
\end{figure}

Here, we extensively present the performance of XTab with reconstruction, contrastive, and supervised pretraining objectives, under light and heavy finetuning. Downstream performance is compared in terms of win rate, model rank, normalized performance, and error reduction rate in Figure \ref{performance_supp}.

\newpage

\section{Identifying the shareable components in XTab} \label{appendix_sharable_component}

\begin{figure}[h]
\vskip 0in
\begin{center}
\centerline{\includegraphics[width=10cm]{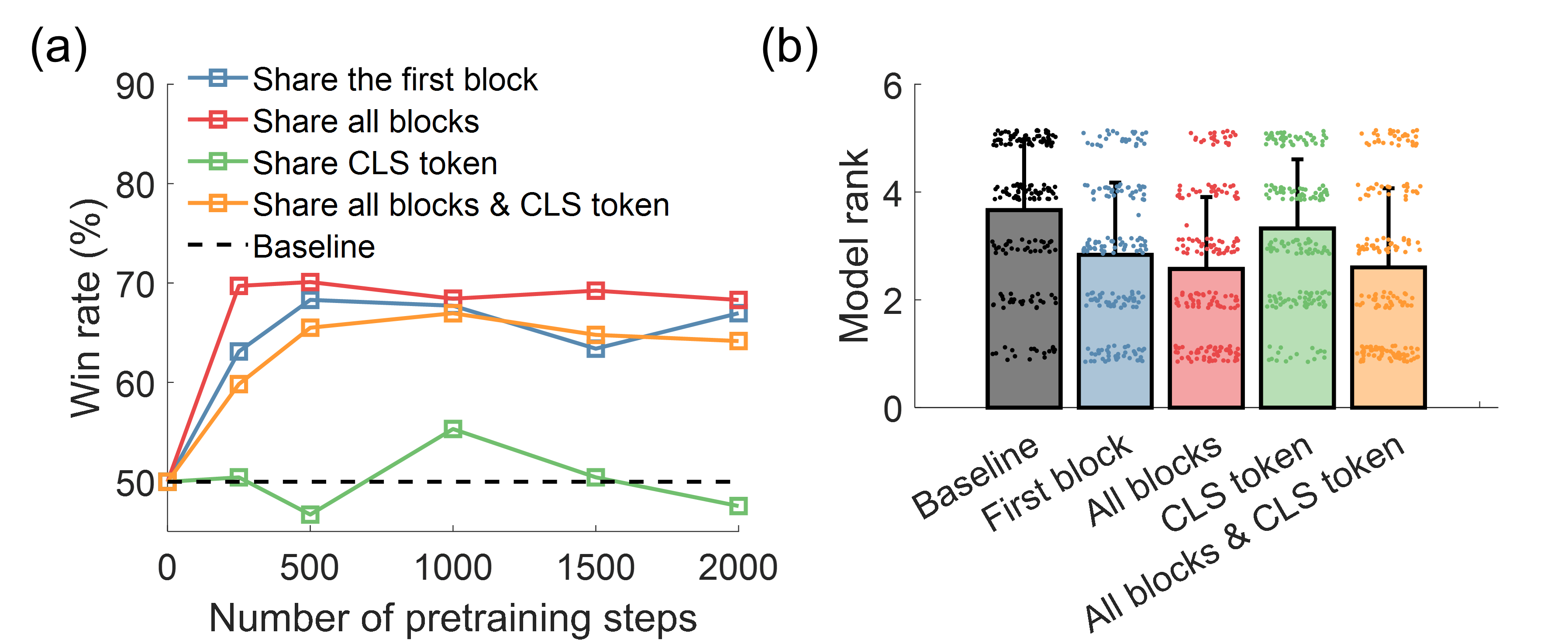}}
\caption{Comparison of XTab with various pretrained components in FT-Transformer. We run this study to understand which component carries general knowledge of tabular tasks and benefits from cross-table pretraining. Several settings are tested, sharing the first block of Transformer, all blocks, $\textsf{[CLS]}$ token, all blocks with $\textsf{[CLS]}$ token, or no component (baseline). Performance is compared in terms of \textbf{(a)} win rate and \textbf{(b)} model rank with light finetuning. Pretraining on the Transformer blocks leads to improved performance, whereas sharing the data-specific $\textsf{[CLS]}$ token is hardly beneficial.}
\label{shareable_blocks}
\end{center}
%\vskip -0.4in
\end{figure}

In XTab, we separate a model into data-specific components (e.g., featurizers and projection heads) and shareable components (Transformer blocks). Only the shareable components are pretrained and contain general knowledge of tabular learning. Therefore, identifying the shareable (or pretrainable) components is critical to the success of cross-table pretraining. In Figure \ref{shareable_blocks}, we run an experiment to pretrain on different FT-Transformer components with the supervised objective. For example, pretraining tasks may share only the first Transformer block and the later two blocks are marked as data-specific. We also let the pretraining tasks share all Transformer blocks, $\textsf{[CLS]}$ token, and all blocks with $\textsf{[CLS]}$ token. As expected, pretraining on the $\textsf{[CLS]}$ token does not lead to improved downstream performance, since $\textsf{[CLS]}$ token is directly related to downstream prediction and thereby highly data-specific. From Figure \ref{shareable_blocks}, we find that it is most beneficial to pretrain on all Transformer blocks without the $\textsf{[CLS]}$ token. Featurizers and projection heads are not shareable since the input/output spaces can be different across tasks.

\section{Finetuning on subsampled datasets}  \label{app_subsample}

\begin{figure}[ht]
\vskip 0.2in
\begin{center}
\centerline{\includegraphics[width=10cm]{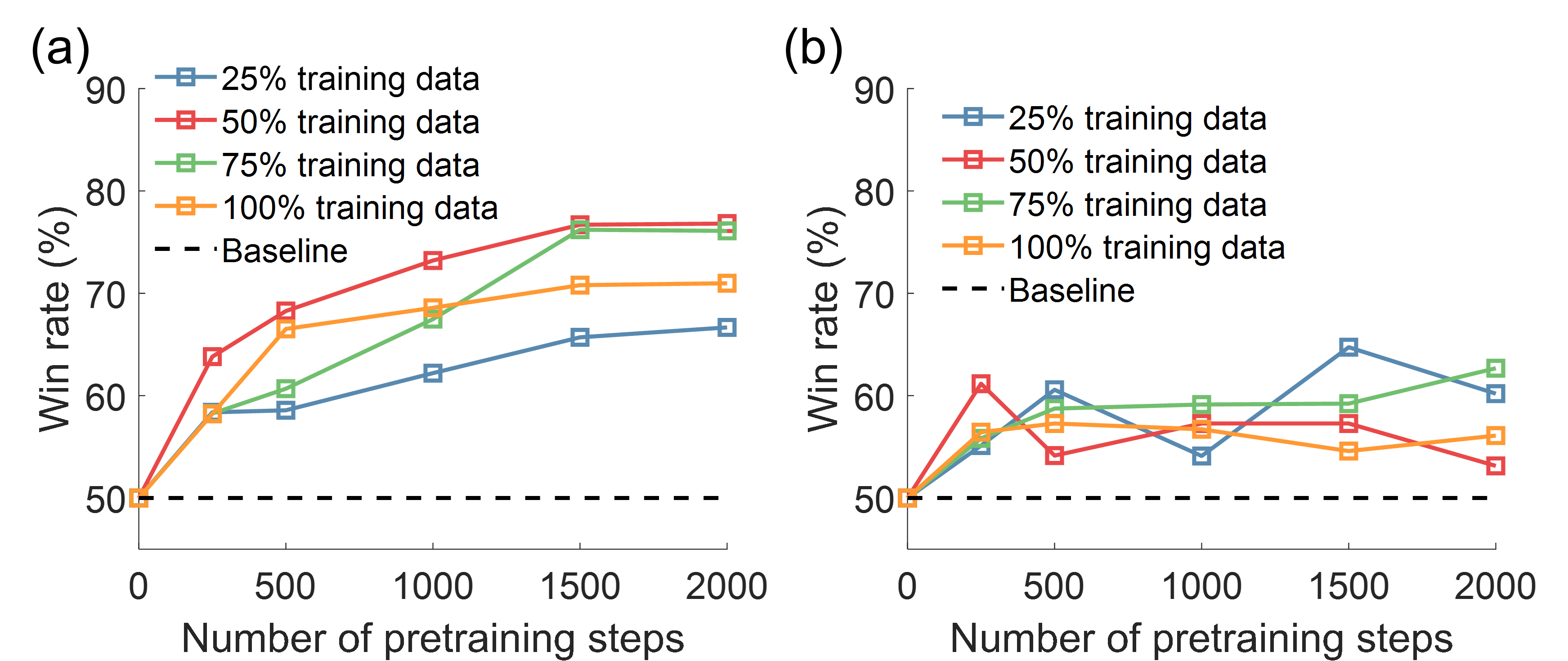}}
\caption{Downstream prediction performance with different sizes of finetuning set. We subsample the rows of tables (i.e., samples) used for finetuning to a fraction of 25\%, 50\%, 75\%, and 100\% (no subsampling). The comparison is performed with \textbf{(a)} light and \textbf{(b)} heavy finetuning. }
\label{few_shot}
\end{center}
\vskip -0.2in
\end{figure}

In addition to light and heavy finetuning, we further tune the pretrained backbone using datasets of different sizes. The backbone is a FT-Transformer model pretrained with the reconstruction objective. We subsampled the training sets of downstream tasks (i.e., finetuning set) by 25\%, 50\%, and 75\%. The finetuning is performed on the reduced datasets to simulate the cases where training data is insufficient. Figure \ref{few_shot} shows the downstream performance with (a) light and (b) heavy finetuning.

All settings in Figure \ref{few_shot} show a clear improvement over the baseline. However, the advantage of XTab does not become more significant with reduced finetuning data. This is partially due to the fact that sufficient finetuning data is still needed to train featurizers and projection heads from scratch. For the same reason, XTab is not compatible with zero-shot learning.

\section{Tuning the size of pretraining set} \label{app_num_tasks}
\begin{figure}[ht]
\vskip 0.2in
\begin{center}
\centerline{\includegraphics[width=10cm]{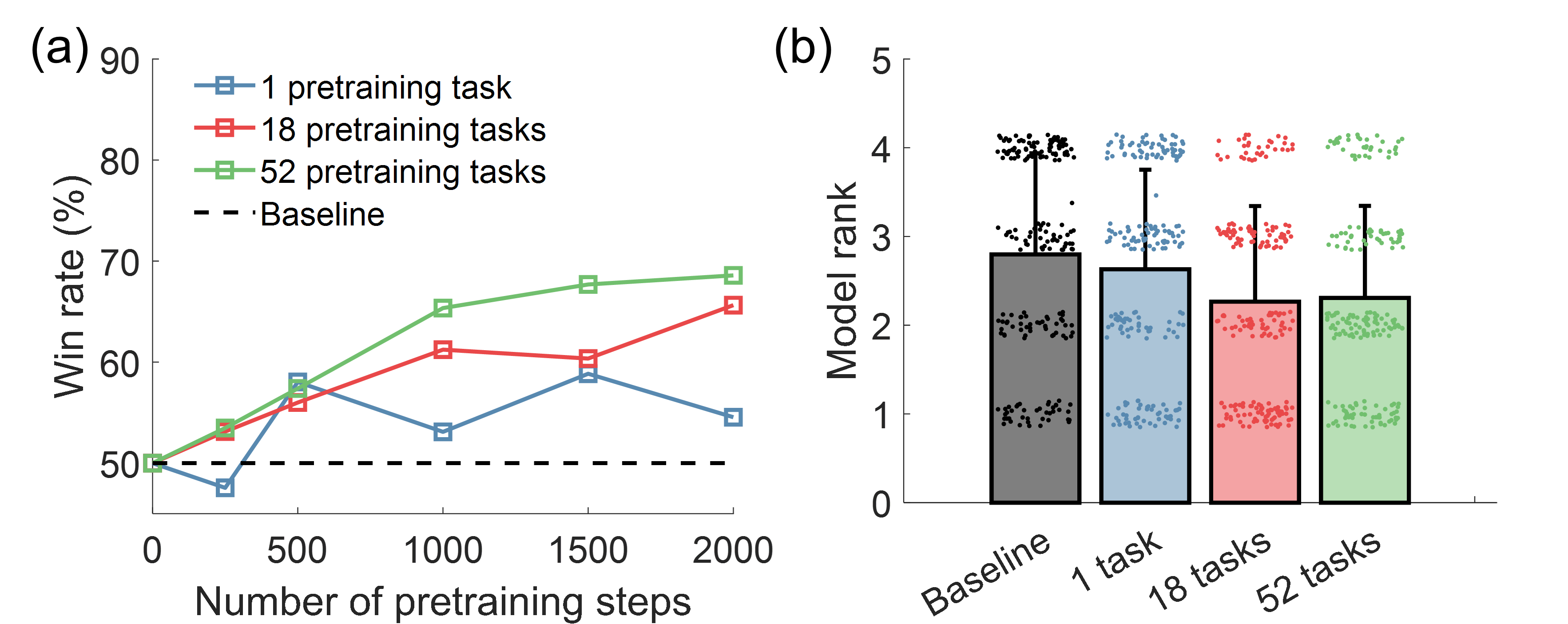}}
\caption{Comparison of XTab pretrained on different numbers of tabular tasks. We pretrain the FT-Transformer backbone using 1 task, 18 tasks and 52 tasks. We compare the downstream prediction performance using \textbf{(a)} win rate and \textbf{(b)} model rank of different approaches. As we use more tasks for pretraining, we observe an improvement in downstream performance.}
\label{num_tasks}
\end{center}
\vskip -0.2in
\end{figure}

The pretrained backbone is expected to host general knowledge that is shared across multiple pretraining tasks. We use different numbers of tabular tasks to pretrain the FT-Transformer using the reconstruction objective. Figure \ref{num_tasks} compares the backbone pretrained on 1 task (Adult income, OpenML task id 359983), 18 tasks, and 52 tasks (selected by the alphabetical order of the task names) with light finetuning. Figure \ref{num_tasks}(a) shows the win rate and Figure \ref{num_tasks}(b) compares the model rank. Figure \ref{num_tasks} indicates that XTab benefits from more pretraining tasks. With many tables involved in cross-table pretraining, XTab can better learn the general knowledge which benefits the downstream performance.

\section{Tuning parameters of federated pretraining} \label{app_fedavg}
\begin{figure}[ht]
\vskip 0.2in
\begin{center}
\centerline{\includegraphics[width=10cm]{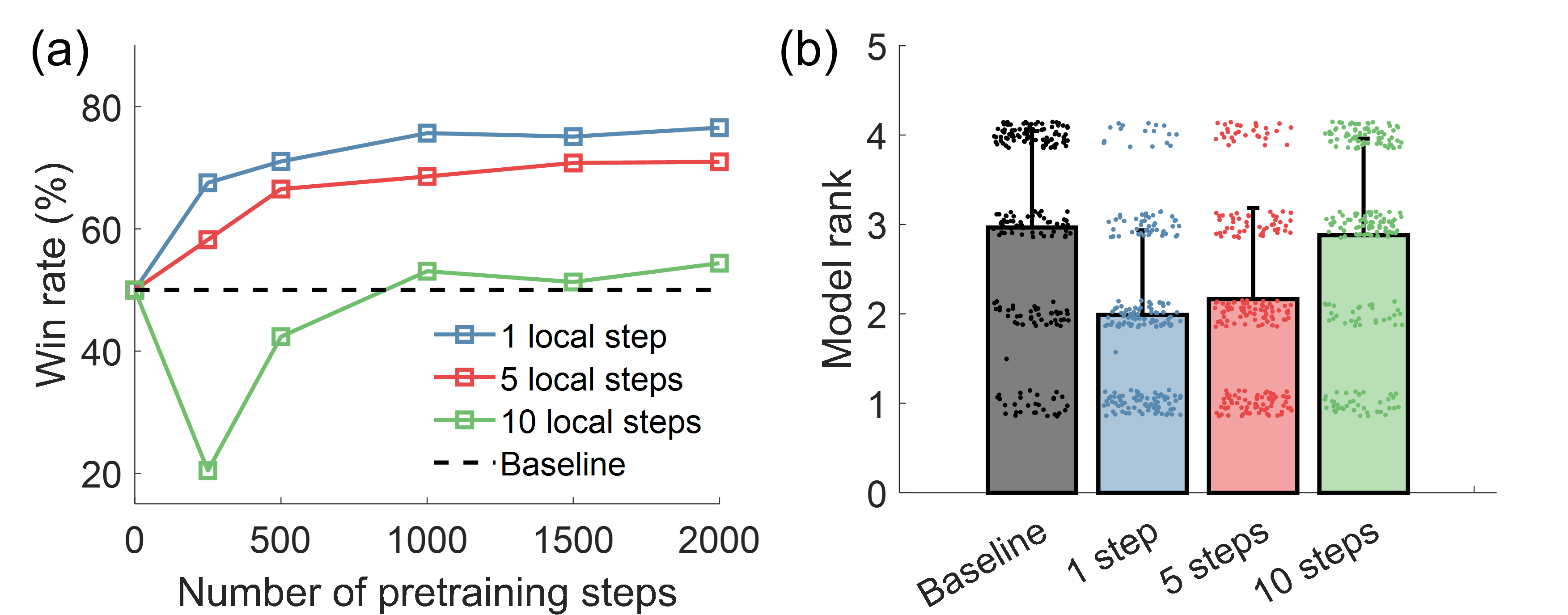}}
\caption{Comparison of federated pretraining settings in XTab. We test FedAvg with different values of $N$, which represents the number of local optimization steps per global aggregation. We compare the downstream prediction performance in terms of \textbf{(a)} win rate and \textbf{(b)} model rank. Both figures suggest that the downstream performance decreases with more local steps in FedAvg. }
\label{FedAvg}
\end{center}
\vskip -0.4in
\end{figure}

XTab uses federated learning to account for a large number of pretraining tasks. We have several clients which perform optimization locally for one task, and a central server that aggregates the gradients from all client nodes. We tune the hyperparameter $N$ in FedAvg (see Section \ref{federated_pretrain}), which indicates the number of local optimization steps between the aggregation steps at the server. We pretrained FT-Transformers with the reconstruction objective and various choices of $N$. Figure \ref{FedAvg} compares the downstream performance with $N = $1, 5, and 10. We notice that the downstream performance decreases as $N$ takes larger numbers. As $N$ increases, there is less communication overhead between the central server and clients. Therefore, we can use $N$ to control the trade-off between the communication cost of federated pretraining and the downstream performance.

\section{Comparison to pretraining without external tasks}
Without external tasks, models are simply pretrained on the downstream training set. Indeed, this is a key difference between XTab and existing tabular pretraining models. SubTab \citep{ucar2021subtab}, SCARF \cite{bahri2021scarf} and SAINT \citep{somepalli2021saint} all use the downstream data for both pretraining and finetuning. Here, we run the experiments to compare XTab against models pretrained without external tasks. We used the “heavy” setting and reconstruction loss. The model details are described as follows:

\begin{itemize} \vspace{-0.1in}
\itemsep-0.3em 
\item w/o external task: random initialization $\rightarrow$ pretrain on downstream task $\rightarrow$ finetune on downstream task.
\item baseline: random initialization $\rightarrow$ finetune on downstream task 
\item w/ external tasks (XTab): XTab initialization (using external tables) $\rightarrow$ pretrain on downstream task $\rightarrow$ finetune on downstream task
\vspace{-0.1in}
\end{itemize}

Here, “w/o external task” is pretrained using the downstream training set. Comparing “w/o external task” and “w/ external task”, the only difference lies in whether we use the XTab-pretrained transformer as initialization, which can indicate the importance of leveraging cross-table information. “Baseline” model does not use pretraining.

\begin{table}[h]
\caption{Comparison to pretraining without external tasks.}
\label{external_tasks}
\vskip 0.15in
\begin{center}
\begin{tabular}{lccc}
\toprule
 & w/o external task & baseline & w/ external tasks \\
\midrule
win rate (against w/o external task)  & 50\% & 35.2\% & 55.7\% \\
\bottomrule
\end{tabular}
\end{center}
\vskip -0.1in
\end{table}

From Table \ref{external_tasks}, we learn that “w/ external task” has a win rate of 55.7\% over “w/ external task”. Pretraining methods generally outperform baseline. This comparison helps illustrate the benefits of XTab in leveraging information across tasks.

\section{Implementation of Saint-v} \label{app_saint_v}

\begin{figure}[ht]
\vskip 0.2in
\begin{center}
\centerline{\includegraphics[width=11cm]{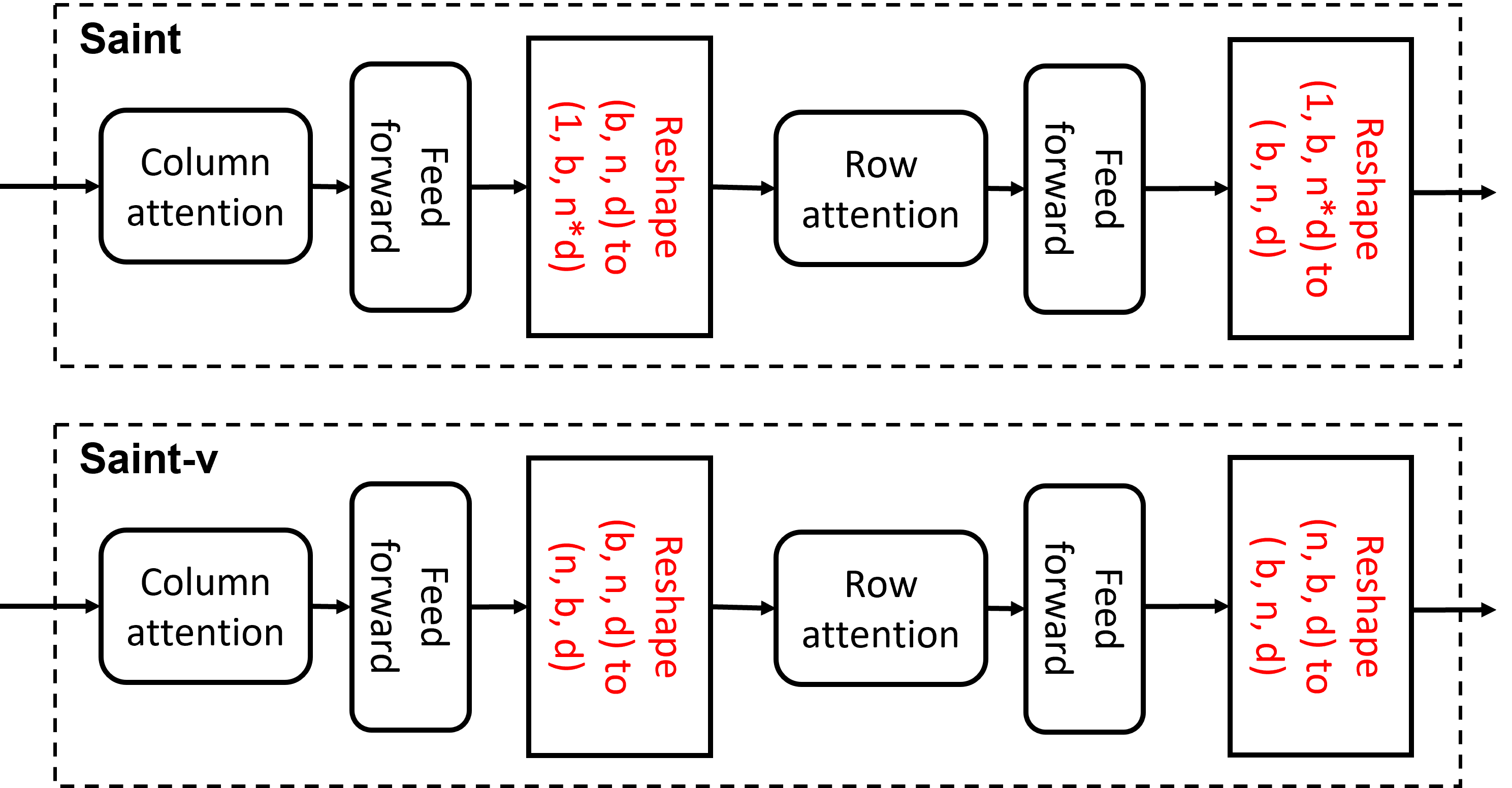}}
\caption{Model structure of Saint and Saint-v. The difference lies in the reshaping operation. Here, $b$ refers to batch size, $n$ is the length of the sequence, and $d$ is the dimension of embedding. The parameter count of Saint is dependent on the number of table columns (i.e., $n$), whereas Saint-v is applicable to all tables with the same structure. }
\label{saint_v}
\end{center}
\vskip -0.2in
\end{figure}

In Figure \ref{saint_v}, we show the difference between the original Saint implementation \citep{somepalli2021saint} and our proposed variation, Saint-v,  to fit into cross-table pretraining. Saint and Saint-v both have a row attention layer to account for the cross-sample interaction. The main difference between Saint and Saint-v lies in the reshaping operation. Saint increases the size of token embeddings by a factor equal to the sequence length. The number of trainable parameters in Saint is dependent on the token count \citep{somepalli2021saint}, making it infeasible for cross-table training. Saint-v transposes the first (batch) and second (number of tokens) dimensions of the input, without altering the dimension of token embeddings. Therefore, Saint-v can be used to process tables with variable columns.

\section{Visualization of pretrained weights}

\begin{figure}[ht]
\vskip 0.2in
\begin{center}
\centerline{\includegraphics[width=13cm]{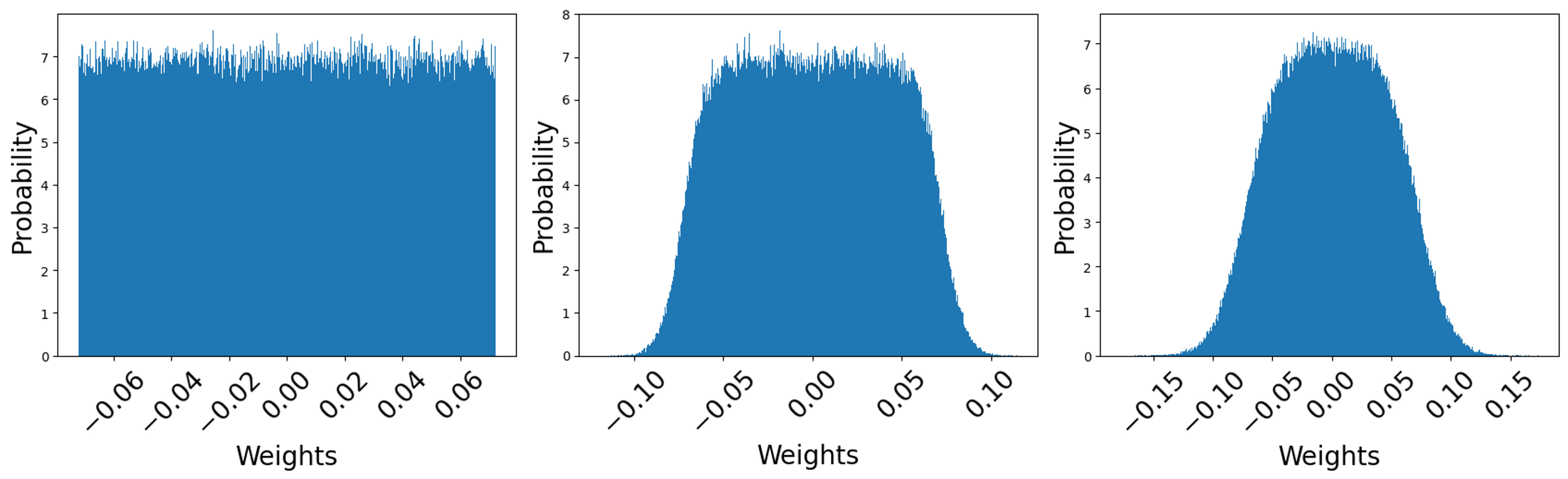}}
\caption{Parameters of FT-Transformer before cross-table pretraining (left), 50 steps after cross-table pretraining (middle), and 500 steps after pretraining (right). The model weights are initialized using a \textit{Kaiming} uniform distribution. With XTab pretraining, the weights converge to a normal distribution. }
\label{weights}
\end{center}
\vskip -0.2in
\end{figure}

To understand the impact of cross-table pretraining on Transformer parameters, we visualize the weight distribution before and after pretraining (Figure \ref{weights}). Here, we ignore the layer normalization and bias terms. Before pretraining, Transformer weights are initialized with \textit{Kaiming} uniform distribution \citep{he2015delving}. The weight distribution converges to a normal distribution with increased pretraining steps. 

\section{Benchmark configurations} \label{benchmark_configurations}

\subsection{Tree-based models}
As tree-based models are known to achieve state-of-the-art performance on tabular tasks \citep{grinsztajn2022tree}, we include popular tree ensemble methods in the benchmark such as XGBoost \citep{chen2016xgboost}, LightGBM \citep{ke2017lightgbm}, CatBoost \citep{dorogush2018catboost}, and Random Forest. Tables \ref{xgb}, \ref{lgbm}, \ref{cat}, and \ref{rf} include the default hyperparameters used for tree-based models and the search space of HPO. We use the default hyperparameters, early stopping strategy, and feature preprocessing logic implemented in AutoGluon 0.5.3 release for each of these models \citep{erickson2020autogluon}, which achieves state-of-the-art performance on AutoML Benchmark \citep{gijsbers2022amlb}. The HPO search space is kept the same as \citet{hollmann2022tabpfn}. 

For gradient-boosted trees (i.e., XGBoost, LightGBM, CatBoost), we apply early stopping to determine the optimal number of boosting rounds (early\_stopping\_rounds = adaptive). Specifically, we use the an early stopping patience of 300 if the training table has less than 10k rows. The patience is reduced by a factor of num\_rows/10k if the row count goes beyond 10k. A minimal early stopping patience of 20 is set to all tables regardless of the table size. 

For Random Forest, we use max\_features to indicate the number of features to consider when making a split. Here max\_features = auto means max\_features=sqrt(n\_features) where n\_features denotes the column count of the training table.

\begin{table}[h]
\caption{XGBoost hyperparameter space.}
\label{xgb}
\vskip 0.15in
\begin{center}
\begin{tabular}{lcccr}
\toprule
Parameter & Default & HPO search space \\
\midrule
learning\_rate  &  0.1 & UniformLog[exp(-7), 1] \\
max\_depth & 6 & UniformInt[1, 10] \\
subsample & 1 & Uniform[0.2, 1]  \\
colsample\_bytree  &  1 &  Uniform[0.2, 1] \\
colsample\_bylevel & 1 &  Uniform[0.2, 1] \\
min\_child\_weight & 1 & UniformLog[exp(-16), exp(5)]  \\
reg\_alpha & 0 & UniformLog[exp(-16), exp(2)]  \\
reg\_lambda & 1 & UniformLog[exp(-16), exp(2)] \\
gamma & 0 & UniformLog[exp(-16), exp(2)]  \\
n\_estimators & 10000 & UniformInt[100, 4000]  \\
booster & gbtree & gbtree \\
early\_stopping\_rounds & adaptive$^{*}$ & adaptive \\
\bottomrule
\multicolumn{3}{l}{\footnotesize{$^*$ The early\_stopping\_rounds depends on the size of data with a minimal patience}}\\
\multicolumn{3}{l}{\footnotesize{ of 20 and maximal patience of 300 rounds.}}\\
\end{tabular}
\end{center}
\vskip -0.1in
\end{table}

\begin{table}[h]
\caption{LighGBM hyperparameter space.}
\label{lgbm}
\vskip 0.15in
\begin{center}
\begin{tabular}{lcccr}
\toprule
Parameter & Default & HPO search space \\
\midrule
num\_leaves  & 31 & UniformInt[5, 50] \\
max\_depth & inf & UniformInt[3, 20] \\
learning\_rate & 0.05 & UniformLog[exp(-3), 1]  \\
n\_estimators  &  10000 & UniformInt[50, 2000] \\
min\_child\_weight & 1e-3 & UniformLog[exp(-5), exp(4)] \\
reg\_alpha & 0 & Categorical[0, 0.1, 1, 2, 5, 7, 10, 50 , 100]  \\
reg\_lambda & 0 & Categorical[0, 0.1, 1, 5, 10, 20, 50, 100] \\
subsample & 1 & Uniform[0.2, 0.8]  \\
early\_stopping\_rounds & adaptive$^*$ & adaptive \\
\bottomrule
\multicolumn{3}{l}{\footnotesize{$^*$ The early\_stopping\_rounds depends on the size of data with a minimal patience}}\\
\multicolumn{3}{l}{\footnotesize{ of 20 and maximal patience of 300 rounds.}}\\
\end{tabular}
\end{center}
\vskip -0.1in
\end{table}

\begin{table}[h]
\caption{CatBoost hyperparameter space.}
\label{cat}
\vskip 0.15in
\begin{center}
\begin{tabular}{lcccr}
\toprule
Parameter & Default & HPO search space \\
\midrule
learning\_rate  &  0.05 & UniformLog[exp(-5), 1] \\
random\_strength & 1 & UniformInt[1, 20] \\
l2\_leaf\_reg & 3 & UniformLog[exp(-3), 1]  \\
bagging\_temperature  &  1 & Uniform[0, 1] \\
leaf\_estimation\_iterations & 1 & UniformInt[1, 20] \\
iterations & 10000 & UniformInt[100, 4000]  \\
early\_stopping\_rounds & adaptive$^*$ & adaptive \\
\bottomrule
\multicolumn{3}{l}{\footnotesize{$^*$ The early\_stopping\_rounds depends on the size of data with a minimal patience}}\\
\multicolumn{3}{l}{\footnotesize{ of 20 and maximal patience of 300 rounds.}}\\
\end{tabular}
\end{center}
\vskip -0.1in
\end{table}

\begin{table}[h]
\caption{Random forest hyperparameter space.}
\label{rf}
\vskip 0.15in
\begin{center}
\begin{tabular}{lcccr}
\toprule
Parameter & Default & HPO search space \\
\midrule
n\_estimators  &  300 & UniformInt[10, 1000] \\
max\_features & auto & Categorical[auto, 0.5, 0.25] \\
max\_leaf\_nodes & inf & UniformInt[100, 4000]  \\

\bottomrule
\end{tabular}
\end{center}
\vskip -0.1in
\end{table}

\subsection{Neural network and FastAI}
We use the tabular neural network from AutoGluon which is implemented on top of PyTorch \citep{erickson2020autogluon}. We use ReLU activation between layers. The default hyperparameters and search space of HPO are listed in Table \ref{nn}.

We also include the FastAI tabular model in this benchmark, which is essentially a neural network that automatically configures the embedding sizes of input features \citep{howard2020fastai}. We use the AutoGluon implementation and default hyperparameters/HPO search spaces suggested by AutoGluon. Detailed configurations of FastAI tabular model is listed in Table \ref{fastai}. 

\begin{table}[h]
\caption{Neural network hyperparameter space.}
\label{nn}
\vskip 0.15in
\begin{center}
\begin{tabular}{lcccr}
\toprule
Parameter & Default & HPO search space \\
\midrule
num\_epochs  &  300 & 300 \\
early\_stop\_patience & 20 & 20 \\
learning\_rate & 3e-4 & UniformLog[1e-4, 0.1] \\
weight\_decay & 1e-6 & UniformLog[1e-12, 0.1] \\
num\_layers & 4 & Categorical[2, 3, 4] \\
hidden\_size & 128 & Categorical[128, 256, 512]   \\
\bottomrule
\end{tabular}
\end{center}
\vskip -0.1in
\end{table}

\begin{table}[h]
\caption{FastAI hyperparameter space.}
\label{fastai}
\vskip 0.15in
\begin{center}
\begin{tabular}{lcccr}
\toprule
Parameter & Default & HPO search space \\
\midrule
num\_epochs  & 30 & Uniform[5, 30] \\
early\_stop\_patience & 20 & 20 \\
learning\_rate & 1e-2 & UniformLog[5e-5, 0.1] \\
weight\_decay & 1e-6 & UniformLog[1e-12, 0.1] \\
\multirow{3}{*}{layers$^*$} & \multirow{3}{*}{none} & \multirow{3}{7cm}{Categorical[none, (200, 100), (200), (500), (1000), (500, 200), (50, 25), (1000, 500), (200, 100, 50), (500, 200, 100), (1000, 500, 200)]} \\
&&\\
&&\\
\bottomrule
\multicolumn{3}{l}{\footnotesize{$^*$ This indicates both the layer count and hidden dimension at each layer.}}
\end{tabular}
\end{center}
\vskip -0.1in
\end{table}

\subsection{TransTab} \label{app_transtab}
We use the official implementation of TransTab v0.0.3 \citep{wang2022transtab}. Since regression tasks are not yet supported by this version, the model rank and training time in Table \ref{t1} are reported only on classification tasks. Specifically, we report the rank of TransTab models to all other methods. For example, if we have the AUC scores of model 1 $>$ TransTab $>$ model 2, then model 1 ranks \#1, model 2 ranks \#2, and TransTab gets a ranking of \#1.5. TransTab rank is \#0.5 with TransTab $>$ model 1 $>$ model 2, and \#2.5 with model 1 $>$ model 2 $>$ TransTab. The inclusion of TransTab in the comparison will not alter the rank of other models, but the rank shows the relative standing of TransTab with respect to other models. Therefore, we can compare the ranking of all methods in Table \ref{t1} even without TransTab regression performance. In Table \ref{t1_class}, we show the regular ranking of TransTab on classification tasks.

The hyperparameters of TransTab is listed in Table \ref{transtab}. We test both the conventional supervised learning setting (TransTab-sl) and the contrastive learning setting which follows the pretraining-finetuning process (TransTab-cl). We use the target-aware contrastive learning objective as it is shown to perform better than its unsupervised counterpart in \citet{wang2022transtab}. Hyperparameters are kept as default whenever possible. We use the column type information from AutoML Benchmark to identify numerical and categorical columns. TransTab-cl performs better than TransTab-sl in our benchmark, as shown in Table \ref{t1}.

\begin{table}[h]
\caption{TransTab hyperparameter for the base and pretraining settings.}
\label{transtab}
\vskip 0.15in
\begin{center}
\begin{tabular}{lcccr}
\toprule
Parameter & supervised learning & contrastive pretraining \\
\midrule

num\_partition & & 4 \\
overlap\_ratio & & 0.5\\
max\_pretrain\_epochs &  & 50 \\
pretrain\_batch\_size &  & 128 \\
pretrain\_learning\_rate &  & 1e-4 \\
max\_epochs & 50 & 50 \\
batch\_size & 238 & 128 \\
learning\_rate & 1e-4 & 1e-4 \\
num\_layers  & 2 & 2 \\
hidden\_dim & 128 & 128 \\
patience & 5 & 5  \\
num\_attention\_heads & 8 & 8  \\
\bottomrule
\end{tabular}
\end{center}
\vskip -0.1in
\end{table}

\subsection{FT-Transformer}
Table \ref{ftt} summarize the general hyperparameters of FT-Transformer. We include three configurations of FT-Transformer in the benchmark:

\textbf{FTT-l:} FT-Transformer with light training. FT-Transformer is trained for maximum 3 epochs. We save the model after each epoch and retrieve the best checkpoint based on the validation performance.  

\textbf{FTT-h:} FT-Transformer with heavy training. FT-Transformer is trained with an early stopping patience of 3. We save the model after each epoch and retrieve the best checkpoint based on the validation performance.

\textbf{FTT-best:} FT-Transformer for the best performance. FT-Transformer is trained with an early stopping patience of 20. We save the model after each 0.5 epoch (i.e., val\_check\_interval = 0.5 in Table \ref{ftt}). At the end of training, we retrieve the best 3 checkpoints based on the validation performance (i.e., top\_k = 3 in Table \ref{ftt}). The checkpoints are averaged using model soup for improved prediction performance \citep{wortsman2022model}.

From FTT-l $\rightarrow$ FTT-h $\rightarrow$ FTT-best, we achieve better tabular prediction performance with increased training time.

\begin{table}[h]
\caption{FT-Transformer hyperparameter space.}
\label{ftt}
\vskip 0.15in
\begin{center}
\begin{tabular}{lcccr}
\toprule
Parameter & Default & HPO search space \\
\midrule
num\_epochs  & inf & inf \\
early\_stop\_patience & 20 & 20 \\
num\_blocks  &  3 & 3 \\
hidden\_size & 192 & 192 \\
num\_attention\_heads & 8 & 8  \\
batch\_size & 128 & Categorical[128, 32, 8, 1]  \\
val\_check\_interval & 1 or 0.5 & Categorical[0.5, 1]  \\
top\_k & 1 or 3 & Categorical[1, 3, 5] \\
\bottomrule
\end{tabular}
\end{center}
\vskip -0.1in
\end{table}

\subsection{XTab}
XTab uses exactly the same structure as FT-Transformer, but with pretrained parameters to initialize the model. Similar to FTT-l/FTT-h/FTT-best, we have XTab-l/XTab-h/XTab-best that follow the same finetuning configurations. We pretrain XTab with the reconstruction loss and FT-Transformer as the backbone. $N=1$ is used for federated pretraining since it achieves the best performance in Figure \ref{FedAvg}. With default hyperparameters, we pretrain the backbone for 2000 rounds, and the number of pretraining iterations is considered as a hyperparameter in HPO. Table \ref{xtab} summarizes the details of XTab.

\begin{table}[h]
\caption{XTab hyperparameter space.}
\label{xtab}
\vskip 0.15in
\begin{center}
\begin{tabular}{lcccr}
\toprule
Parameter & Default & HPO search space \\
\midrule
\multicolumn{3}{c}{\textbf{All default parameters and search spaces from FT-Transformer}} \\
N\_FedAvg & 1 & 1 \\
pretrain\_objective & reconstruction & reconstruction \\
num\_pretrain\_rounds & 2000 & Categorical[0, 250, 1000, 2000] \\
\bottomrule
\end{tabular}
\end{center}
\vskip -0.1in
\end{table}

\begin{table}[ht]
\caption{This table is similar to Table \ref{t1}, but compares the tabular models on 48 classification tasks. Since TransTab v0.0.3 does not supports regression tasks, we include this table for classification tasks only.}
\label{t1_class}
\vskip 0.15in
\begin{center}
\begin{tabular}{llccr}
\toprule
& Methods & Time (s) & Rank  \\
\midrule
\multirow{13}{*}{ \rotatebox{90}{Default hyperparameter} }
& RF & 11.39  & 7.58 $\pm$ 4.19 \\
& XGBoost & 11.90 & 5.10 $\pm$ 3.41\\
& LightGBM & 8.62 & 5.58 $\pm$ 3.54 \\
& \textbf{CatBoost} & \textbf{229.36} & \textbf{3.02 $\pm$ 2.87} \\
\cmidrule{2-4}
& FastAI & 27.01 & 7.27 $\pm$ 3.79 \\
& NN & 73.64 &  6.96 $\pm$ 3.66\\
\cmidrule{2-4}
& TransTab-sl & 342.49 &  12.33 $\pm$ 2.68\\
& TransTab-cl & 331.98 &  11.60 $\pm$ 3.13\\
\cmidrule{2-4}
& FTT-l & 74.91 & 10.94 $\pm$ 2.54 \\
& XTab-l & 74.48 & 10.06 $\pm$ 2.88 \\
\cmidrule{2-4}
& FTT-h & 309.64 & 7.23 $\pm$ 2.17 \\
& XTab-h & 291.19 & 7.35 $\pm$ 1.92 \\
\cmidrule{2-4}
& FTT-best & 544.77 &  5.33 $\pm$ 2.43 \\
& \textbf{XTab-best} & \textbf{472.35} &  \textbf{4.63 $\pm$ 2.28} \\
\midrule
\end{tabular}
\end{center}
\vskip -0.1in
\end{table}

\section{Dataset statistics}
Table \ref{ds_stat} shows the statistics of all datasets from the AutoML Benchmark \citep{gijsbers2022amlb}, including the task name, type, and table dimensions. We equally split the benchmark into 2 folds for pretraining and downstream evaluation. Therefore, there is minimal overlap between pretraining tasks and downstream tasks. The success of XTab in this setting demonstrates the ability of learning general knowledge across all downstream tasks. 

\begin{table}[!ht]
\caption{Dataset statistics of AutoML Benchmark. We split the benchmark into 2 folds. We use fold 1 to pretrain XTab and fold 2 to evaluate downstream performance, and vice versa. 20 out of the 104 datasets failed during our experiments. They are marked with symbols and excluded from the comparison.}
    
    \centering
    \label{ds_stat}
    \scalebox{0.6}{
    \begin{tabular}{l|lllll|llll}
    \toprule
        ~ & name & num\_rows & num\_columns & task\_type & ~ & name & num\_rows & num\_columns & task\_type \\ \hline
        \multirow{52}{*}{ \rotatebox{90}{Fold 1} } 
        & APSFailure & 76000 & 171 & binary & \multirow{52}{*}{ \rotatebox{90}{Fold 2} }  & dna & 3186 & 181 & multiclass \\ 
        ~ & Airlines\_DepDelay\_10M & 10000000 & 10 & regression & ~ & elevators & 16599 & 19 & regression \\ 
        ~ & Allstate\_Claims\_Severity & 188318 & 131 & regression & ~ & eucalyptus & 736 & 20 & multiclass \\ 
        ~ & Amazon\_employee\_access & 32769 & 10 & binary & ~ & fabert$^*$ & 8237 & 801 & multiclass \\ 
        ~ & Australian & 690 & 15 & binary & ~ & first-order-theorem-proving & 6118 & 52 & multiclass \\ 
        ~ & Bioresponse$^*$ & 3751 & 1777 & binary & ~ & gina$^*$ & 3153 & 971 & binary \\ 
        ~ & Brazilian\_houses & 10692 & 13 & regression & ~ & guillermo$^*$ & 20000 & 4297 & binary \\ 
        ~ & Buzzinsocialmedia\_Twitter & 583250 & 78 & regression & ~ & helena & 65196 & 28 & multiclass \\ 
        ~ & Click\_prediction\_small & 39948 & 12 & binary & ~ & house\_16H & 22784 & 17 & regression \\ 
        ~ & Diabetes130US & 101766 & 50 & multiclass & ~ & house\_prices\_nominal & 1460 & 80 & regression \\ 
        ~ & Fashion-MNIST$^*$ & 70000 & 785 & multiclass & ~ & house\_sales & 21613 & 22 & regression \\ 
        ~ & GesturePhaseSegmentationProcessed & 9873 & 33 & multiclass & ~ & jannis & 83733 & 55 & multiclass \\ 
        ~ & Higgs & 1000000 & 29 & binary & ~ & jasmine & 2984 & 145 & binary \\ 
        ~ & Internet-Advertisements$^*$ & 3279 & 1559 & binary & ~ & jungle\_chess\_2pcs\_raw\_endgame\_complete & 44819 & 7 & multiclass \\ 
        ~ & KDDCup09-Upselling$^*$ & 50000 & 14892 & binary & ~ & kc1 & 2109 & 22 & binary \\ 
        ~ & KDDCup09\_appetency & 50000 & 231 & binary & ~ & kick & 72983 & 33 & binary \\ 
        ~ & KDDCup99$^\dagger$ & 4898431 & 42 & multiclass & ~ & kr-vs-kp & 3196 & 37 & binary \\ 
        ~ & MIP-2016-regression & 1090 & 145 & regression & ~ & madeline & 3140 & 260 & binary \\ 
        ~ & Mercedes\_Benz\_Greener\_Manufacturing & 4209 & 377 & regression & ~ & mfeat-factors & 2000 & 217 & multiclass \\ 
        ~ & MiniBooNE & 130064 & 51 & binary & ~ & micro-mass$^*$ & 571 & 1301 & multiclass \\ 
        ~ & Moneyball & 1232 & 15 & regression & ~ & nomao & 34465 & 119 & binary \\ 
        ~ & OnlineNewsPopularity & 39644 & 60 & regression & ~ & numerai28\_6 & 96320 & 22 & binary \\ 
        ~ & PhishingWebsites & 11055 & 31 & binary & ~ & nyc-taxi-green-dec-2016 & 581835 & 19 & regression \\ 
        ~ & QSAR-TID-10980$^*$ & 5766 & 1026 & regression & ~ & okcupid-stem & 50789 & 20 & multiclass \\ 
        ~ & QSAR-TID-11$^*$ & 5742 & 1026 & regression & ~ & ozone-level-8hr & 2534 & 73 & binary \\ 
        ~ & SAT11-HAND-runtime-regression & 4440 & 117 & regression & ~ & pc4 & 1458 & 38 & binary \\ 
        ~ & Santander\_transaction\_value$^*$ & 4459 & 4992 & regression & ~ & philippine & 5832 & 309 & binary \\ 
        ~ & Satellite & 5100 & 37 & binary & ~ & phoneme & 5404 & 6 & binary \\ 
        ~ & Yolanda & 400000 & 101 & regression & ~ & pol & 15000 & 49 & regression \\ 
        ~ & abalone & 4177 & 9 & regression & ~ & porto-seguro & 595212 & 58 & binary \\ 
        ~ & ada & 4147 & 49 & binary & ~ & qsar-biodeg & 1055 & 42 & binary \\ 
        ~ & adult & 48842 & 15 & binary & ~ & quake & 2178 & 4 & regression \\ 
        ~ & airlines & 539383 & 8 & binary & ~ & riccardo$^*$ & 20000 & 4297 & binary \\ 
        ~ & albert & 425240 & 79 & binary & ~ & robert$^*$ & 10000 & 7201 & multiclass \\ 
        ~ & amazon-commerce-reviews$^*$ & 1500 & 10001 & multiclass & ~ & segment & 2310 & 20 & multiclass \\ 
        ~ & arcene$^*$ & 100 & 10001 & binary & ~ & sensory & 576 & 12 & regression \\ 
        ~ & bank-marketing & 45211 & 17 & binary & ~ & sf-police-incidents & 2215023 & 9 & binary \\ 
        ~ & black\_friday & 166821 & 10 & regression & ~ & shuttle & 58000 & 10 & multiclass \\ 
        ~ & blood-transfusion-service-center & 748 & 5 & binary & ~ & socmob & 1156 & 6 & regression \\ 
        ~ & boston & 506 & 14 & regression & ~ & space\_ga & 3107 & 7 & regression \\ 
        ~ & car & 1728 & 7 & multiclass & ~ & steel-plates-fault & 1941 & 28 & multiclass \\ 
        ~ & christine$^*$ & 5418 & 1637 & binary & ~ & sylvine & 5124 & 21 & binary \\ 
        ~ & churn & 5000 & 21 & binary & ~ & tecator & 240 & 125 & regression \\ 
        ~ & cmc & 1473 & 10 & multiclass & ~ & topo\_2\_1 & 8885 & 267 & regression \\ 
        ~ & cnae-9$^*$ & 1080 & 857 & multiclass & ~ & us\_crime & 1994 & 127 & regression \\ 
        ~ & colleges & 7063 & 45 & regression & ~ & vehicle & 846 & 19 & multiclass \\ 
        ~ & connect-4 & 67557 & 43 & multiclass & ~ & volkert & 58310 & 181 & multiclass \\ 
        ~ & covertype & 581012 & 55 & multiclass & ~ & wilt & 4839 & 6 & binary \\ 
        ~ & credit-g & 1000 & 21 & binary & ~ & wine-quality-white & 4898 & 12 & multiclass \\ 
        ~ & diamonds & 53940 & 10 & regression & ~ & wine\_quality & 6497 & 12 & regression \\ 
        ~ & dilbert$^*$ & 10000 & 2001 & multiclass & ~ & yeast & 1484 & 9 & multiclass \\ 
        ~ & dionis$^{\dagger\dagger}$ & 416188 & 61 & multiclass & ~ & yprop\_4\_1 & 8885 & 252 & regression \\ 
        \bottomrule
    \multicolumn{6}{l}{\footnotesize{$^*$ Out of memory error for FT-Transformers and XTab with a batch size of 128.}}\\
    \multicolumn{6}{l}{\footnotesize{$^\dagger$ Timeout error for FT-Transformers and XTab with a 1-hour training time budget.}}\\
    \multicolumn{6}{l}{\footnotesize{$^{\dagger\dagger}$ Out of memory error for Random Forest.}}\\
    \end{tabular}
    }
\end{table}

\section{Raw prediction performance} \label{raw_prediction_performance}
Here, we present the raw prediction performance on AutoML Benchmark in Table \ref{t11}, \ref{t12} and \ref{t13}. Please refer to Table \ref{t1} for the aggregated comparison. 20 datasets are excluded from the benchmark since they fail to fit into the 16 GB GPU memory. We report the performance on the remaining 84 downstream tasks. All experiments are repeated for 5 trials and we report the average performance.

\begin{table}[!ht]
    \centering
    \label{t11}
    \caption{Raw prediction performance on AutoML Benchmark of the following models: Random Forest (RF), XGBoost, LightGBM, CatBoost, tabular neural network from AutoGluon (NN), FastAI tabular model, and TransTab with contrastive pretraining (TransTab-cl). All models use the default hyperparameters as specified in Appendix \ref{benchmark_configurations}. We use AUC scores as the evaluation metric for binary classification ($\uparrow$), log loss for multicloass classification ($\downarrow$) and RMSE for regression tasks ($\downarrow$). Regression tasks are not supported by TransTab v0.0.3 by the time this experiment was conducted. Zoom in for better view.}
    \scalebox{0.5}{
    \begin{tabular}{llllllllll}
    \toprule
        name & task type & metrics & RF & XGB & LGBM & CAT & FastAI & NN & TransTab-cl \\ \hline
        APSFailure & binary & AUC & 0.9901 & 0.9917 & 0.992 & 0.9932 & 0.9803 & 0.9901 & 0.9815 \\ 
        Amazon\_employee\_access & binary & AUC & 0.8534 & 0.8416 & 0.8541 & 0.8989 & 0.8315 & 0.8289 & 0.7606 \\ 
        Australian & binary & AUC & 0.9328 & 0.9237 & 0.9273 & 0.9396 & 0.9314 & 0.9284 & 0.8825 \\ 
        Click\_prediction\_small & binary & AUC & 0.6593 & 0.7012 & 0.6968 & 0.7067 & 0.6539 & 0.6876 & 0.6583 \\ 
        Higgs & binary & AUC & 0.815 & 0.8321 & 0.8337 & 0.8364 & 0.8454 & 0.8438 & 0.6864 \\ 
        KDDCup09\_appetency & binary & AUC & 0.774 & 0.826 & 0.7967 & 0.8404 & 0.729 & 0.8042 & NaN \\ 
        MiniBooNE & binary & AUC & 0.9807 & 0.9857 & 0.9856 & 0.9862 & 0.9418 & 0.9868 & 0.8047 \\ 
        PhishingWebsites & binary & AUC & 0.9955 & 0.9967 & 0.997 & 0.9961 & 0.9966 & 0.9959 & 0.8215 \\ 
        Satellite & binary & AUC & 0.977 & 0.9475 & 0.9342 & 0.9725 & 0.9903 & 0.9945 & 0.9832 \\ 
        ada & binary & AUC & 0.9096 & 0.9239 & 0.9206 & 0.9278 & 0.9003 & 0.9124 & 0.9223 \\ 
        adult & binary & AUC & 0.9075 & 0.9282 & 0.9286 & 0.9287 & 0.9122 & 0.9092 & 0.9122 \\ 
        airlines & binary & AUC & 0.721 & 0.7283 & 0.725 & 0.7279 & 0.7204 & 0.7172 & 0.7096 \\ 
        albert & binary & AUC & 0.7362 & 0.7661 & 0.7711 & 0.7853 & 0.7572 & 0.7499 & NaN \\ 
        bank-marketing & binary & AUC & 0.9313 & 0.9364 & 0.9372 & 0.9387 & 0.9369 & 0.9323 & 0.9172 \\ 
        blood-transfusion-service-center & binary & AUC & 0.7245 & 0.7437 & 0.7445 & 0.758 & 0.7726 & 0.7449 & 0.772 \\ 
        churn & binary & AUC & 0.9088 & 0.9203 & 0.92 & 0.9198 & 0.92 & 0.9018 & 0.8081 \\ 
        credit-g & binary & AUC & 0.7882 & 0.743 & 0.7421 & 0.76 & 0.7394 & 0.7441 & 0.7649 \\ 
        jasmine & binary & AUC & 0.8879 & 0.8671 & 0.8703 & 0.8831 & 0.8482 & 0.8501 & 0.8089 \\ 
        kc1 & binary & AUC & 0.8207 & 0.8063 & 0.7952 & 0.8116 & 0.7973 & 0.8012 & 0.7912 \\ 
        kick & binary & AUC & 0.7626 & 0.7822 & 0.7684 & 0.7864 & 0.7674 & 0.765 & 0.6943 \\ 
        kr-vs-kp & binary & AUC & 0.9994 & 0.9995 & 0.9997 & 0.9998 & 0.9996 & 0.9996 & 0.6036 \\ 
        madeline & binary & AUC & 0.8725 & 0.9199 & 0.9233 & 0.9319 & 0.6327 & 0.674 & 0.5966 \\ 
        nomao & binary & AUC & 0.9944 & 0.9961 & 0.9962 & 0.9963 & 0.9918 & 0.9918 & 0.9868 \\ 
        numerai28\_6 & binary & AUC & 0.5153 & 0.5221 & 0.5265 & 0.5296 & 0.5289 & 0.5255 & 0.5287 \\ 
        ozone-level-8hr & binary & AUC & 0.9324 & 0.9234 & 0.923 & 0.9344 & 0.905 & 0.9361 & 0.9072 \\ 
        pc4 & binary & AUC & 0.9377 & 0.9478 & 0.9507 & 0.9519 & 0.9302 & 0.9429 & 0.872 \\ 
        philippine & binary & AUC & 0.8428 & 0.8532 & 0.8637 & 0.8523 & 0.7817 & 0.7781 & 0.7996 \\ 
        phoneme & binary & AUC & 0.9596 & 0.952 & 0.952 & 0.9533 & 0.9326 & 0.9399 & 0.8254 \\ 
        porto-seguro & binary & AUC & 0.6095 & 0.6378 & 0.6285 & 0.6391 & 0.6338 & 0.6292 & NaN \\ 
        qsar-biodeg & binary & AUC & 0.917 & 0.9206 & 0.9199 & 0.9304 & 0.9209 & 0.9258 & 0.9087 \\ 
        sf-police-incidents & binary & AUC & 0.6885 & 0.6766 & 0.6784 & 0.7186 & 0.6051 & 0.6307 & NaN \\ 
        sylvine & binary & AUC & 0.9828 & 0.9844 & 0.9843 & 0.9868 & 0.976 & 0.9717 & 0.965 \\ 
        wilt & binary & AUC & 0.9869 & 0.9885 & 0.9837 & 0.988 & 0.9919 & 0.9725 & 0.9138 \\ 
        Diabetes130US & multiclass & log loss & 0.8555 & 0.8421 & 0.8563 & 0.836 & 0.8703 & 0.8746 & 0.8744 \\ 
        GesturePhaseSegmentationProcessed & multiclass & log loss & 0.8676 & 0.8567 & 0.8513 & 0.8033 & 1.0617 & 1.039 & 1.3868 \\ 
        car & multiclass & log loss & 0.0374 & 0.0186 & 0.0308 & 0.0564 & 0.3106 & 0.0286 & 0.5011 \\ 
        cmc & multiclass & log loss & 0.9991 & 0.9313 & 0.9329 & 0.9118 & 0.9393 & 0.9132 & 1.0093 \\ 
        connect-4 & multiclass & log loss & 0.4858 & 0.3408 & 0.3324 & 0.3681 & 0.3319 & 0.3552 & 0.8451 \\ 
        covertype & multiclass & log loss & 0.1763 & 0.0861 & 0.0924 & 0.1492 & 0.1988 & 0.1452 & NaN \\ 
        dna & multiclass & log loss & 1.1122 & 0.115 & 0.1124 & 0.1135 & 0.187 & 0.1764 & 1.0137 \\ 
        eucalyptus & multiclass & log loss & 0.7148 & 0.797 & 0.7823 & 0.7099 & 0.6963 & 0.6883 & 0.8627 \\ 
        first-order-theorem-proving & multiclass & log loss & 1.1789 & 1.1005 & 1.0987 & 1.0826 & 1.2193 & 1.233 & 1.589 \\ 
        helena & multiclass & log loss & 3.0947 & 2.6571 & 2.7992 & 2.5489 & 2.5741 & 2.5421 & 6.2857 \\ 
        jannis & multiclass & log loss & 0.7196 & 0.6838 & 0.6868 & 0.6771 & 0.6752 & 0.6921 & 0.7557 \\ 
        jungle\_chess\_2pcs\_raw\_endgame\_complete & multiclass & log loss & 0.4116 & 0.2099 & 0.2219 & 0.259 & 0.2342 & 0.1286 & 0.2116 \\ 
        mfeat-factors & multiclass & log loss & 0.1217 & 0.1646 & 0.1523 & 0.1053 & 0.1069 & 0.1023 & 5.391 \\ 
        okcupid-stem & multiclass & log loss & 0.5976 & 0.57 & 0.5722 & 0.564 & 0.5819 & 0.5833 & 0.5824 \\ 
        segment & multiclass & log loss & 0.0797 & 0.0666 & 0.0702 & 0.0514 & 0.0962 & 0.0875 & 0.3558 \\ 
        shuttle & multiclass & log loss & 0.0008 & 0.0005 & 0.0342 & 0.0005 & 0.0094 & 0.0026 & NaN \\ 
        steel-plates-fault & multiclass & log loss & 0.5332 & 0.4905 & 0.4978 & 0.4777 & 0.6685 & 0.5894 & 0.7929 \\ 
        vehicle & multiclass & log loss & 0.4963 & 0.5314 & 0.5178 & 0.5026 & 0.3725 & 0.3798 & 1.1012 \\ 
        volkert & multiclass & log loss & 0.9342 & 0.834 & 0.8418 & 0.7931 & 0.822 & 0.909 & 1.2693 \\ 
        wine-quality-white & multiclass & log loss & 0.8196 & 0.8556 & 0.8822 & 0.8486 & 0.9746 & 0.9719 & 1.2884 \\ 
        yeast & multiclass & log loss & 1.112 & 1.0567 & 1.1105 & 1.0051 & 1.0918 & 1.0397 & 1.2425 \\ 
        Airlines\_DepDelay\_10M & regression & RMSE & 28.9112 & 28.6239 & 28.5986 & 28.6857 & 28.7381 & 30.1015 & NaN \\ 
        Allstate\_Claims\_Severity & regression & RMSE & 1965.848 & 1908.768 & 1900.444 & 1868.448 & 2003.79 & 2014.408 & NaN \\ 
        Brazilian\_houses & regression & RMSE & 5002.8084 & 4451.8282 & 10291.733 & 6976.8256 & 20174.666 & 4011.5132 & NaN \\ 
        Buzzinsocialmedia\_Twitter & regression & RMSE & 179.2258 & 239.4952 & 208.6456 & 256.8753 & 220.5236 & 214.1682 & NaN \\ 
        MIP-2016-regression & regression & RMSE & 764.5954 & 799.144 & 773.5334 & 1326.5238 & 5581.654 & 23970.42 & NaN \\ 
        Mercedes\_Benz\_Greener\_Manufacturing & regression & RMSE & 9.3286 & 8.7689 & 8.813 & 8.614 & 9.5226 & 8.7805 & NaN \\ 
        Moneyball & regression & RMSE & 24.8672 & 23.5196 & 23.823 & 22.809 & 21.9525 & 23.4829 & NaN \\ 
        OnlineNewsPopularity & regression & RMSE & 11843.722 & 11673.084 & 11420.656 & 11383.204 & 11399.604 & 11502.598 & NaN \\ 
        SAT11-HAND-runtime-regression & regression & RMSE & 1148.728 & 1089.708 & 964.2244 & 1116.356 & 1086.366 & 1309.33 & NaN \\ 
        Yolanda & regression & RMSE & 9.1417 & 8.7697 & 8.6998 & 8.6813 & 8.5693 & 8.8721 & NaN \\ 
        abalone & regression & RMSE & 2.1384 & 2.2099 & 2.2091 & 2.1944 & 2.1211 & 2.1677 & NaN \\ 
        black\_friday & regression & RMSE & 3663.638 & 3459.81 & 3452.612 & 3462.058 & 3592.534 & 3717.344 & NaN \\ 
        boston & regression & RMSE & 3.2711 & 3.159 & 3.3991 & 2.6787 & 4.1064 & 3.2801 & NaN \\ 
        colleges & regression & RMSE & 0.1456 & 0.1422 & 0.1398 & 0.1401 & 0.1571 & 0.1569 & NaN \\ 
        diamonds & regression & RMSE & 545.0098 & 540.029 & 525.7748 & 514.932 & 599.311 & 627.6254 & NaN \\ 
        elevators & regression & RMSE & 0.0027 & 0.0021 & 0.0021 & 0.002 & 0.0022 & 0.002 & NaN \\ 
        house\_16H & regression & RMSE & 30202.46 & 28623.1 & 28700.92 & 28230.98 & 29523.65 & 28660.04 & NaN \\ 
        house\_prices\_nominal & regression & RMSE & 26002.08 & 24473.34 & 25573.2667 & 21413.3 & 24193.24 & 25424.98 & NaN \\ 
        house\_sales & regression & RMSE & 122271.4 & 114646.5 & 109766 & 105759.42 & 113428.4 & 143064 & NaN \\ 
        nyc-taxi-green-dec-2016 & regression & RMSE & 1.6163 & 1.8043 & 1.6599 & 1.6454 & 1.7925 & 1.8583 & NaN \\ 
        pol & regression & RMSE & 4.9681 & 4.8782 & 4.4412 & 4.3646 & 3.7933 & 49.9791 & NaN \\ 
        quake & regression & RMSE & 0.1924 & 0.1869 & 0.1851 & 0.1833 & 0.1853 & 0.1862 & NaN \\ 
        sensory & regression & RMSE & 0.6857 & 0.7267 & 0.6847 & 0.6834 & 0.7237 & 0.7533 & NaN \\ 
        socmob & regression & RMSE & 17.5107 & 13.9014 & 12.431 & 11.673 & 14.7385 & 14.4491 & NaN \\ 
        space\_ga & regression & RMSE & 0.1099 & 0.1049 & 0.1017 & 0.1014 & 0.1013 & 0.1016 & NaN \\ 
        tecator & regression & RMSE & 1.3789 & 1.2681 & 1.914 & 1.831 & 1.756 & 1.6347 & NaN \\ 
        topo\_2\_1 & regression & RMSE & 0.0302 & 0.0305 & 0.03 & 0.03 & 0.0306 & 0.0324 & NaN \\ 
        us\_crime & regression & RMSE & 0.1391 & 0.1399 & 0.134 & 0.1347 & 0.1424 & 0.141 & NaN \\ 
        wine\_quality & regression & RMSE & 0.6089 & 0.6211 & 0.6227 & 0.6186 & 0.6767 & 0.712 & NaN \\ 
        yprop\_4\_1 & regression & RMSE & 0.0297 & 0.03 & 0.0299 & 0.0298 & 0.0311 & 0.0344 & NaN \\ 
        \bottomrule
    \end{tabular}
    }
\end{table}

\begin{table}[!ht]
    \caption{Raw prediction performance on AutoML Benchmark of the following models: FT-Transformer with light finetuning (FTT-l), XTab with light finetuning (XTab-l), FT-Transformer with heavy finetuning (FTT-h), XTab with heavy finetuning (XTab-h), FT-Transformer with model soup (FTT-best), and XTab with model soup (XTab-best). All models use the default hyperparameters as specified in Appendix \ref{benchmark_configurations}. We use AUC scores as the evaluation metric for binary classification ($\uparrow$), log loss for multicloass classification ($\downarrow$) and RMSE for regression tasks ($\downarrow$). Zoom in for better view.}
    \centering
    \label{t12}
    \scalebox{0.5}{
    \begin{tabular}{lllllllll}
    \toprule
        name & task type & metrics & FTT-l & XTab-l & FTT-h & XTab-h & FTT-best & XTab-best \\ \hline
        APSFailure & binary & AUC & 0.9889 & 0.9896 & 0.988 & 0.9868 & 0.9859 & 0.9873 \\ 
        Amazon\_employee\_access & binary & AUC & 0.7221 & 0.7454 & 0.7894 & 0.7877 & 0.7952 & 0.7941 \\ 
        Australian & binary & AUC & 0.9036 & 0.9229 & 0.8994 & 0.921 & 0.9197 & 0.921 \\ 
        Click\_prediction\_small & binary & AUC & 0.6711 & 0.6724 & 0.6767 & 0.6752 & 0.6755 & 0.6761 \\ 
        Higgs & binary & AUC & 0.8311 & 0.8327 & 0.8451 & 0.8447 & 0.8473 & 0.8475 \\ 
        KDDCup09\_appetency & binary & AUC & 0.8178 & 0.8205 & 0.8144 & 0.8192 & 0.8152 & 0.8251 \\ 
        MiniBooNE & binary & AUC & 0.9664 & 0.9663 & 0.9778 & 0.9758 & 0.9825 & 0.9813 \\ 
        PhishingWebsites & binary & AUC & 0.9871 & 0.9879 & 0.9936 & 0.9936 & 0.996 & 0.9957 \\ 
        Satellite & binary & AUC & 0.979 & 0.981 & 0.9784 & 0.9822 & 0.9928 & 0.9854 \\ 
        ada & binary & AUC & 0.9058 & 0.9109 & 0.9148 & 0.9169 & 0.9202 & 0.9194 \\ 
        adult & binary & AUC & 0.9142 & 0.9153 & 0.9148 & 0.9148 & 0.916 & 0.9161 \\ 
        airlines & binary & AUC & 0.7064 & 0.7082 & 0.7136 & 0.7132 & 0.7153 & 0.7151 \\ 
        albert & binary & AUC & 0.7478 & 0.7507 & 0.7552 & 0.7551 & 0.7562 & 0.7561 \\ 
        bank-marketing & binary & AUC & 0.9283 & 0.9342 & 0.9382 & 0.9376 & 0.9403 & 0.939 \\ 
        blood-transfusion-service-center & binary & AUC & 0.7636 & 0.7582 & 0.7615 & 0.7498 & 0.7625 & 0.751 \\ 
        churn & binary & AUC & 0.888 & 0.8794 & 0.9127 & 0.9044 & 0.9157 & 0.916 \\ 
        credit-g & binary & AUC & 0.7448 & 0.7299 & 0.7587 & 0.7485 & 0.7442 & 0.747 \\ 
        jasmine & binary & AUC & 0.8399 & 0.8449 & 0.8556 & 0.8595 & 0.8614 & 0.8692 \\ 
        kc1 & binary & AUC & 0.7998 & 0.7915 & 0.7998 & 0.7939 & 0.8001 & 0.8035 \\ 
        kick & binary & AUC & 0.7717 & 0.774 & 0.7752 & 0.7739 & 0.7766 & 0.7771 \\ 
        kr-vs-kp & binary & AUC & 0.9773 & 0.9892 & 0.9984 & 0.9991 & 0.9993 & 0.9998 \\ 
        madeline & binary & AUC & 0.5902 & 0.6034 & 0.708 & 0.8393 & 0.8548 & 0.8869 \\ 
        nomao & binary & AUC & 0.9882 & 0.9902 & 0.9919 & 0.9928 & 0.9933 & 0.9937 \\ 
        numerai28\_6 & binary & AUC & 0.5293 & 0.5298 & 0.5287 & 0.5284 & 0.5261 & 0.5283 \\ 
        ozone-level-8hr & binary & AUC & 0.8803 & 0.906 & 0.9322 & 0.9299 & 0.9273 & 0.9329 \\ 
        pc4 & binary & AUC & 0.8688 & 0.8868 & 0.9383 & 0.9451 & 0.9438 & 0.9451 \\ 
        philippine & binary & AUC & 0.757 & 0.7765 & 0.7988 & 0.8158 & 0.823 & 0.8315 \\ 
        phoneme & binary & AUC & 0.8968 & 0.9136 & 0.9165 & 0.9256 & 0.9468 & 0.9432 \\ 
        porto-seguro & binary & AUC & 0.636 & 0.6364 & 0.6351 & 0.6351 & 0.6368 & 0.6373 \\ 
        qsar-biodeg & binary & AUC & 0.8861 & 0.8773 & 0.9113 & 0.9087 & 0.9181 & 0.9189 \\ 
        sf-police-incidents & binary & AUC & 0.6131 & 0.6129 & 0.6048 & 0.6037 & 0.6068 & 0.607 \\ 
        sylvine & binary & AUC & 0.9669 & 0.971 & 0.981 & 0.98 & 0.9817 & 0.9861 \\ 
        wilt & binary & AUC & 0.989 & 0.992 & 0.9893 & 0.988 & 0.9903 & 0.9888 \\ 
        Diabetes130US & multiclass & log loss & 0.8575 & 0.8538 & 0.8468 & 0.8472 & 0.8426 & 0.8455 \\ 
        GesturePhaseSegmentationProcessed & multiclass & log loss & 1.2019 & 1.1886 & 1.0364 & 1.0555 & 0.9685 & 1.0197 \\ 
        car & multiclass & log loss & 0.3607 & 0.355 & 0.0616 & 0.0611 & 0.0023 & 0.0004 \\ 
        cmc & multiclass & log loss & 0.9795 & 0.9688 & 0.9735 & 0.9362 & 0.9591 & 0.9398 \\ 
        connect-4 & multiclass & log loss & 0.5482 & 0.4899 & 0.3592 & 0.353 & 0.3383 & 0.3332 \\ 
        covertype & multiclass & log loss & 0.2743 & 0.266 & 0.1463 & 0.146 & 0.1333 & 0.1332 \\ 
        dna & multiclass & log loss & 0.8681 & 0.3408 & 0.1761 & 0.1337 & 0.1429 & 0.1292 \\ 
        eucalyptus & multiclass & log loss & 1.0905 & 1.2154 & 0.7786 & 0.7435 & 0.7387 & 0.7056 \\ 
        first-order-theorem-proving & multiclass & log loss & 1.4326 & 1.3986 & 1.269 & 1.2362 & 1.2199 & 1.1937 \\ 
        helena & multiclass & log loss & 2.8484 & 2.8462 & 2.5574 & 2.5552 & 2.5496 & 2.5399 \\ 
        jannis & multiclass & log loss & 0.7123 & 0.7015 & 0.6689 & 0.672 & 0.6655 & 0.6646 \\ 
        jungle\_chess\_2pcs\_raw\_endgame\_complete & multiclass & log loss & 0.2817 & 0.2781 & 0.022 & 0.0202 & 0.0107 & 0.0106 \\ 
        mfeat-factors & multiclass & log loss & 1.6934 & 1.5505 & 0.1439 & 0.1352 & 0.1227 & 0.114 \\ 
        okcupid-stem & multiclass & log loss & 0.5723 & 0.5717 & 0.5715 & 0.5746 & 0.5694 & 0.5701 \\ 
        segment & multiclass & log loss & 0.335 & 0.2667 & 0.1169 & 0.1189 & 0.0772 & 0.0788 \\ 
        shuttle & multiclass & log loss & 0.0018 & 0.0021 & 0.0022 & 0.0023 & 0.0014 & 0.0017 \\ 
        steel-plates-fault & multiclass & log loss & 0.9308 & 0.9095 & 0.5837 & 0.5857 & 0.5649 & 0.5424 \\ 
        vehicle & multiclass & log loss & 0.9964 & 1.0895 & 0.4769 & 0.4469 & 0.4325 & 0.405 \\ 
        volkert & multiclass & log loss & 1.1074 & 1.0797 & 0.8092 & 0.8105 & 0.7847 & 0.8046 \\ 
        wine-quality-white & multiclass & log loss & 1.047 & 1.0441 & 1.0143 & 0.99 & 0.9883 & 0.9861 \\ 
        yeast & multiclass & log loss & 1.2193 & 1.226 & 1.0339 & 1.0373 & 1.0156 & 1.016 \\ 
        Airlines\_DepDelay\_10M & regression & RMSE & 28.7656 & 28.7608 & 28.7771 & 28.7766 & 28.7682 & 28.8381 \\ 
        Allstate\_Claims\_Severity & regression & RMSE & 1916.358 & 1907.124 & 1902.972 & 1897.556 & 1885.78 & 1881.712 \\ 
        Brazilian\_houses & regression & RMSE & 9132.3466 & 11103.2593 & 8243.249 & 8453.9666 & 8132.8652 & 8729.3638 \\ 
        Buzzinsocialmedia\_Twitter & regression & RMSE & 206.7792 & 208.0826 & 170.2322 & 166.302 & 160.4322 & 161.9 \\ 
        MIP-2016-regression & regression & RMSE & 26528.74 & 25235.92 & 4605.84 & 1890.9452 & 1052.837 & 882.3568 \\ 
        Mercedes\_Benz\_Greener\_Manufacturing & regression & RMSE & 10.3715 & 9.3503 & 8.9875 & 8.8223 & 8.688 & 8.6548 \\ 
        Moneyball & regression & RMSE & 32.4144 & 29.7766 & 23.2309 & 22.5419 & 21.7374 & 21.8931 \\ 
        OnlineNewsPopularity & regression & RMSE & 11361.304 & 11360.136 & 11365.064 & 11347.134 & 11353.516 & 11346.508 \\ 
        SAT11-HAND-runtime-regression & regression & RMSE & 1751.088 & 1584.554 & 1602.846 & 1276.914 & 1060.4908 & 1040.6616 \\ 
        Yolanda & regression & RMSE & 8.8256 & 8.7725 & 8.7038 & 8.6963 & 8.6265 & 8.6506 \\ 
        abalone & regression & RMSE & 2.272 & 2.18 & 2.2423 & 2.1597 & 2.1565 & 2.1381 \\ 
        black\_friday & regression & RMSE & 3536.97 & 3530.13 & 3522.2775 & 3523.254 & 3500.544 & 3497.502 \\ 
        boston & regression & RMSE & 6.7548 & 6.5448 & 3.9548 & 3.8535 & 3.7662 & 2.9211 \\ 
        colleges & regression & RMSE & 0.1587 & 0.1557 & 0.1555 & 0.1504 & 0.1456 & 0.1466 \\ 
        diamonds & regression & RMSE & 575.2152 & 557.6404 & 558.863 & 560.7662 & 519.0348 & 520.1262 \\ 
        elevators & regression & RMSE & 0.0021 & 0.002 & 0.002 & 0.002 & 0.0019 & 0.0019 \\ 
        house\_16H & regression & RMSE & 33217.86 & 31728.76 & 30478.9 & 31508.2 & 28847.02 & 29216.04 \\ 
        house\_prices\_nominal & regression & RMSE & 42374.86 & 35212.56 & 26234.8 & 23914.88 & 22393.1 & 21866.12 \\ 
        house\_sales & regression & RMSE & 120387 & 126072.8 & 117748 & 117384.8 & 110948.4 & 112808.6 \\ 
        nyc-taxi-green-dec-2016 & regression & RMSE & 1.8388 & 1.8233 & 1.8209 & 1.7333 & 1.7446 & 1.6899 \\ 
        pol & regression & RMSE & 8.8125 & 5.7178 & 2.9935 & 3.078 & 2.1899 & 2.1846 \\ 
        quake & regression & RMSE & 0.1843 & 0.1834 & 0.1833 & 0.1835 & 0.1836 & 0.1851 \\ 
        sensory & regression & RMSE & 0.7746 & 0.7556 & 0.7498 & 0.7494 & 0.7475 & 0.7817 \\ 
        socmob & regression & RMSE & 20.9773 & 19.2464 & 19.1815 & 19.192 & 19.0985 & 19.1424 \\ 
        space\_ga & regression & RMSE & 0.1257 & 0.1215 & 0.1126 & 0.1103 & 0.1034 & 0.1018 \\ 
        tecator & regression & RMSE & 12.8959 & 12.7553 & 6.5291 & 5.4309 & 2.7824 & 1.6988 \\ 
        topo\_2\_1 & regression & RMSE & 0.0306 & 0.0304 & 0.0304 & 0.0303 & 0.0302 & 0.0301 \\ 
        us\_crime & regression & RMSE & 0.157 & 0.1471 & 0.1386 & 0.1382 & 0.1352 & 0.1352 \\ 
        wine\_quality & regression & RMSE & 0.7117 & 0.7066 & 0.7021 & 0.701 & 0.6812 & 0.6801 \\ 
        yprop\_4\_1 & regression & RMSE & 0.0304 & 0.0303 & 0.0303 & 0.0303 & 0.0303 & 0.0302 \\ 
        \bottomrule
    \end{tabular}
    }
\end{table}

\begin{table}[!ht]
    \caption{Raw prediction performance on AutoML Benchmark under the HPO setting. All models use the HPO search spaces as specified in Appendix \ref{benchmark_configurations}.}
    \centering
    \label{t13}
    \scalebox{0.6}{
    \begin{tabular}{lllllllllll}
    \toprule
        name & task\_type & metrics & RF & XGB & LGBM & CAT & FastAI & NN & FTT & XTab \\ \hline
        APSFailure & binary & AUC & 0.9891 & 0.9929 & 0.9905 & 0.9923 & 0.9825 & 0.9896 & 0.9859 & 0.9875 \\ 
        Amazon\_employee\_access & binary & AUC & 0.8629 & 0.8526 & 0.8555 & 0.8995 & 0.8535 & 0.8329 & 0.7945 & 0.7929 \\ 
        Australian & binary & AUC & 0.9331 & 0.9382 & 0.9399 & 0.9362 & 0.9272 & 0.9211 & 0.9184 & 0.9132 \\ 
        Click\_prediction\_small & binary & AUC & 0.6976 & 0.7017 & 0.6953 & 0.7105 & 0.681 & 0.6964 & 0.675 & 0.6757 \\ 
        Higgs & binary & AUC & 0.8126 & 0.8365 & 0.8345 & 0.8367 & 0.8485 & 0.8435 & 0.8458 & 0.8329 \\ 
        KDDCup09\_appetency & binary & AUC & 0.8186 & 0.8307 & 0.8041 & 0.8367 & 0.762 & 0.8168 & 0.8159 & 0.8127 \\ 
        MiniBooNE & binary & AUC & 0.9813 & 0.9866 & 0.9863 & 0.9865 & 0.9845 & 0.9878 & 0.9823 & 0.9799 \\ 
        PhishingWebsites & binary & AUC & 0.9964 & 0.997 & 0.9966 & 0.9961 & 0.9965 & 0.9968 & 0.9961 & 0.9961 \\ 
        Satellite & binary & AUC & 0.9746 & 0.9443 & 0.9821 & 0.9873 & 0.9935 & 0.9945 & 0.9908 & 0.9879 \\ 
        ada & binary & AUC & 0.9227 & 0.9237 & 0.9215 & 0.9247 & 0.9055 & 0.9175 & 0.9197 & 0.9185 \\ 
        adult & binary & AUC & 0.9176 & 0.9288 & 0.928 & 0.929 & 0.9143 & 0.9138 & 0.9154 & 0.9167 \\ 
        airlines & binary & AUC & 0.7252 & 0.7301 & 0.7262 & 0.7266 & 0.7204 & 0.7192 & 0.7154 & 0.7128 \\ 
        albert & binary & AUC & 0.7342 & 0.7687 & 0.7758 & 0.7846 & 0.7569 & 0.7653 & 0.7559 & 0.7499 \\ 
        bank-marketing & binary & AUC & 0.9318 & 0.9364 & 0.9385 & 0.9388 & 0.9367 & 0.9354 & 0.9411 & 0.9405 \\ 
        blood-transfusion-service-center & binary & AUC & 0.7273 & 0.7166 & 0.7503 & 0.759 & 0.7443 & 0.7227 & 0.7451 & 0.7303 \\ 
        churn & binary & AUC & 0.907 & 0.9089 & 0.9131 & 0.9194 & 0.9192 & 0.9156 & 0.914 & 0.9168 \\ 
        credit-g & binary & AUC & 0.791 & 0.7512 & 0.7498 & 0.7779 & 0.7527 & 0.7458 & 0.7481 & 0.743 \\ 
        jasmine & binary & AUC & 0.8875 & 0.875 & 0.8596 & 0.873 & 0.8516 & 0.8542 & 0.8606 & 0.8579 \\ 
        kc1 & binary & AUC & 0.8163 & 0.8154 & 0.7904 & 0.8069 & 0.7972 & 0.7984 & 0.7979 & 0.8062 \\ 
        kick & binary & AUC & 0.7699 & 0.7855 & 0.7708 & 0.786 & 0.7771 & 0.7735 & 0.7773 & 0.7775 \\ 
        kr-vs-kp & binary & AUC & 0.9998 & 0.9988 & 0.9997 & 0.9997 & 0.9985 & 0.9995 & 0.9989 & 0.9998 \\ 
        madeline & binary & AUC & 0.9275 & 0.9364 & 0.9176 & 0.938 & 0.7825 & 0.7752 & 0.8628 & 0.8923 \\ 
        nomao & binary & AUC & 0.9946 & 0.9963 & 0.9961 & 0.996 & 0.9928 & 0.9923 & 0.9933 & 0.9937 \\ 
        numerai28\_6 & binary & AUC & 0.5277 & 0.5243 & 0.5262 & 0.5263 & 0.5282 & 0.5258 & 0.5258 & 0.5266 \\ 
        ozone-level-8hr & binary & AUC & 0.9303 & 0.9231 & 0.9259 & 0.9307 & 0.9256 & 0.9446 & 0.9277 & 0.9293 \\ 
        pc4 & binary & AUC & 0.9459 & 0.9366 & 0.9437 & 0.9425 & 0.9415 & 0.9397 & 0.9412 & 0.9433 \\ 
        philippine & binary & AUC & 0.8498 & 0.8627 & 0.8487 & 0.8541 & 0.7934 & 0.802 & 0.8246 & 0.8324 \\ 
        phoneme & binary & AUC & 0.9604 & 0.9563 & 0.9521 & 0.9573 & 0.9332 & 0.9428 & 0.9539 & 0.9532 \\ 
        porto-seguro & binary & AUC & 0.63 & 0.6419 & 0.6345 & 0.6394 & 0.6358 & 0.634 & 0.6369 & 0.6362 \\ 
        qsar-biodeg & binary & AUC & 0.9162 & 0.9091 & 0.9146 & 0.9031 & 0.9187 & 0.9181 & 0.9196 & 0.9174 \\ 
        sf-police-incidents & binary & AUC & 0.6706 & 0.686 & 0.681 & 0.7158 & 0.6122 & 0.6474 & 0.6068 & 0.607 \\ 
        sylvine & binary & AUC & 0.9838 & 0.9863 & 0.985 & 0.9866 & 0.9826 & 0.9811 & 0.9846 & 0.9846 \\ 
        wilt & binary & AUC & 0.9877 & 0.9901 & 0.991 & 0.9811 & 0.9808 & 0.9898 & 0.9898 & 0.9941 \\ 
        Diabetes130US & multiclass & log loss & 0.8519 & 0.8357 & 0.8499 & 0.8355 & 0.8643 & 0.8665 & 0.8433 & 0.8489 \\ 
        GesturePhaseSegmentationProcessed & multiclass & log loss & 0.8598 & 0.8242 & 0.8328 & 0.7833 & 1.0472 & 0.9798 & 0.9604 & 0.9604 \\ 
        car & multiclass & log loss & 0.0504 & 0.3288 & 0.2972 & 0.0578 & 0.2856 & 0.0013 & 0.0002 & 0 \\ 
        cmc & multiclass & log loss & 0.9074 & 0.9305 & 0.9117 & 0.9237 & 0.9387 & 0.9264 & 0.9519 & 0.9449 \\ 
        connect-4 & multiclass & log loss & 0.497 & 0.3269 & 0.3218 & 0.3719 & 0.3215 & 0.3373 & 0.3383 & 0.3537 \\ 
        covertype & multiclass & log loss & 0.1824 & 0.0889 & 0.0915 & 0.109 & 0.1346 & 0.1264 & 0.1373 & 0.2386 \\ 
        dna & multiclass & log loss & 0.1487 & 0.0989 & 0.1102 & 0.1182 & 0.1484 & 0.1489 & 0.1279 & 0.131 \\ 
        eucalyptus & multiclass & log loss & 0.7119 & 0.7358 & 0.7493 & 0.7476 & 0.7189 & 0.7247 & 0.7481 & 0.7305 \\ 
        first-order-theorem-proving & multiclass & log loss & 1.0671 & 1.0664 & 1.0849 & 1.0858 & 1.2051 & 1.1899 & 1.212 & 1.1831 \\ 
        helena & multiclass & log loss & 2.7036 & 2.5968 & 2.6022 & 2.5647 & 2.5305 & 2.513 & 2.5355 & 2.5407 \\ 
        jannis & multiclass & log loss & 0.7072 & 0.6731 & 0.6807 & 0.6764 & 0.6694 & 0.6555 & 0.662 & 0.6603 \\ 
        jungle\_chess\_2pcs\_raw\_endgame\_complete & multiclass & log loss & 0.3169 & 0.2299 & 0.2257 & 0.2335 & 0.2097 & 0.0475 & 0.012 & 0.0122 \\ 
        mfeat-factors & multiclass & log loss & 0.1636 & 0.1201 & 0.1382 & 0.1114 & 0.1089 & 0.0773 & 0.1099 & 0.1094 \\ 
        okcupid-stem & multiclass & log loss & 0.5902 & 0.5663 & 0.5701 & 0.5637 & 0.5739 & 0.5694 & 0.5688 & 0.5694 \\ 
        segment & multiclass & log loss & 0.0762 & 0.0718 & 0.0714 & 0.067 & 0.0905 & 0.0818 & 0.0812 & 0.0932 \\ 
        shuttle & multiclass & log loss & 0.0006 & 0.0004 & 0.0005 & 0.0005 & 0.0077 & 0.0028 & 0.0013 & 0.0013 \\ 
        steel-plates-fault & multiclass & log loss & 0.5287 & 0.4937 & 0.4912 & 0.4834 & 0.6348 & 0.5823 & 0.568 & 0.5536 \\ 
        vehicle & multiclass & log loss & 0.4972 & 0.4555 & 0.5123 & 0.5383 & 0.3649 & 0.4504 & 0.4303 & 0.4256 \\ 
        volkert & multiclass & log loss & 0.9181 & 0.8078 & 0.8199 & 0.7951 & 0.801 & 0.8266 & 0.7847 & 0.8004 \\ 
        wine-quality-white & multiclass & log loss & 0.803 & 0.793 & 0.8602 & 0.8198 & 0.9771 & 0.9703 & 0.9789 & 0.9708 \\ 
        yeast & multiclass & log loss & 1.02 & 1.0213 & 1.0999 & 1.0018 & 1.054 & 1.0349 & 1.0156 & 1.0155 \\ 
        Airlines\_DepDelay\_10M & regression & RMSE & 28.9108 & 28.577 & 28.5797 & 28.7851 & 28.7342 & 30.1429 & 28.7435 & 28.809 \\ 
        Allstate\_Claims\_Severity & regression & RMSE & 1939.89 & 1887.014 & 1885.37 & 1866.698 & 1977.002 & 1892.888 & 1885.936 & 1905.3 \\ 
        Brazilian\_houses & regression & RMSE & 5285.2022 & 4488.908 & 8505.7592 & 9491.7438 & 16486.544 & 3859.9434 & 8264.7402 & 8201.4656 \\ 
        Buzzinsocialmedia\_Twitter & regression & RMSE & 179.265 & 241.4524 & 200.1286 & 229.5252 & 168.3526 & 177.9844 & 162.3476 & 173.2894 \\ 
        MIP-2016-regression & regression & RMSE & 765.0452 & 800.3702 & 829.5368 & 823.6524 & 2377.88 & 3903.15 & 871.013 & 878.175 \\ 
        Mercedes\_Benz\_Greener\_Manufacturing & regression & RMSE & 8.9261 & 8.6234 & 8.7048 & 8.6512 & 9.0556 & 8.7859 & 8.6845 & 8.7014 \\ 
        Moneyball & regression & RMSE & 24.4026 & 23.0216 & 24.5429 & 22.8522 & 22.0157 & 23.1796 & 21.5883 & 21.8534 \\ 
        OnlineNewsPopularity & regression & RMSE & 11464.464 & 11364.592 & 11397.174 & 11410.652 & 11378.526 & 11478.684 & 11379.368 & 11365.422 \\ 
        SAT11-HAND-runtime-regression & regression & RMSE & 1139.12 & 1067.37 & 968.7284 & 1100.046 & 1079.6472 & 1166.976 & 1034.3146 & 1032.3848 \\ 
        Yolanda & regression & RMSE & 9.229 & 8.6079 & 8.7664 & 8.701 & 8.6134 & 8.7159 & 8.6318 & 8.7462 \\ 
        abalone & regression & RMSE & 2.1789 & 2.1927 & 2.2062 & 2.2103 & 2.1402 & 2.1496 & 2.1335 & 2.142 \\ 
        black\_friday & regression & RMSE & 3503.918 & 3452.056 & 3452.454 & 3463.792 & 3573.808 & 3592.846 & 3500.544 & 3513.162 \\ 
        boston & regression & RMSE & 3.3039 & 3.0809 & 3.3606 & 2.945 & 3.3487 & 3.4282 & 3.4638 & 2.8631 \\ 
        colleges & regression & RMSE & 0.1426 & 0.1381 & 0.1407 & 0.1397 & 0.1537 & 0.1529 & 0.147 & 0.1451 \\ 
        diamonds & regression & RMSE & 544.454 & 534.047 & 521.6772 & 517.6136 & 593.0908 & 549.5522 & 520.3338 & 517.9442 \\ 
        elevators & regression & RMSE & 0.0027 & 0.0022 & 0.0021 & 0.002 & 0.0019 & 0.0019 & 0.0018 & 0.0019 \\ 
        house\_16H & regression & RMSE & 29691.38 & 28688.28 & 28892.28 & 27962.54 & 30915.52 & 29078.48 & 27869.3 & 29179 \\ 
        house\_prices\_nominal & regression & RMSE & 25655.78 & 21950.74 & 22964.88 & 21954.12 & 22389.62 & 23721.84 & 22199.78 & 22056.98 \\ 
        house\_sales & regression & RMSE & 121712.2 & 111883.4 & 110022.38 & 107470.58 & 111026.4 & 118402 & 110166.28 & 109626.2 \\ 
        nyc-taxi-green-dec-2016 & regression & RMSE & 1.631 & 1.7843 & 1.6683 & 1.6148 & 1.5909 & 1.737 & 1.7596 & 1.698 \\ 
        pol & regression & RMSE & 4.6848 & 4.6111 & 4.4196 & 3.9429 & 3.6529 & 3.6789 & 2.0737 & 2.0981 \\ 
        quake & regression & RMSE & 0.1845 & 0.1896 & 0.1872 & 0.1851 & 0.1851 & 0.1825 & 0.1843 & 0.1844 \\ 
        sensory & regression & RMSE & 0.6731 & 0.7238 & 0.6924 & 0.6966 & 0.6588 & 0.7263 & 0.7803 & 0.8059 \\ 
        socmob & regression & RMSE & 16.2576 & 12.8328 & 11.3572 & 13.45 & 8.4355 & 11.0054 & 19.1915 & 19.1915 \\ 
        space\_ga & regression & RMSE & 0.1096 & 0.1036 & 0.1035 & 0.1028 & 0.0992 & 0.0982 & 0.1031 & 0.1007 \\ 
        tecator & regression & RMSE & 1.3897 & 0.9691 & 1.1218 & 1.6591 & 1.7807 & 1.6622 & 1.7329 & 1.2897 \\ 
        topo\_2\_1 & regression & RMSE & 0.0302 & 0.03 & 0.0301 & 0.0301 & 0.0302 & 0.0306 & 0.0302 & 0.0301 \\ 
        us\_crime & regression & RMSE & 0.1379 & 0.1343 & 0.1372 & 0.1354 & 0.1391 & 0.1392 & 0.1351 & 0.1351 \\ 
        wine\_quality & regression & RMSE & 0.6004 & 0.6046 & 0.6261 & 0.5972 & 0.6767 & 0.6864 & 0.682 & 0.6761 \\ 
        yprop\_4\_1 & regression & RMSE & 0.0295 & 0.0334 & 0.0298 & 0.0296 & 0.0791 & 0.0303 & 0.0303 & 0.0301 \\ 
        \bottomrule
    \end{tabular}
    }
\end{table}

\end{document}